\documentclass[letterpaper, 10 pt, journal, twoside]{ieeetran} 
\usepackage{graphicx}
\usepackage{amsmath}
\usepackage{bbding}
\usepackage{pifont}
\usepackage{wasysym}
\usepackage{amssymb}
\usepackage{hyperref}
\usepackage{threeparttable} 
\usepackage{cite} 
\usepackage{booktabs}
\usepackage{xcolor}
\usepackage{multirow}
\usepackage{blindtext}
\usepackage{bm}
\usepackage{subcaption}
\usepackage[linesnumbered, ruled]{algorithm2e}

\SetKwInput{KwInput}{Input}
\SetKwInput{KwOutput}{Output}

\usepackage{cite}

\IEEEoverridecommandlockouts                              
\usepackage{geometry}
\usepackage{afterpage}
\usepackage{mwe}
\title{Topo-boundary: A Benchmark Dataset on Topological Road-boundary Detection Using Aerial Images for Autonomous Driving}
\author{Zhenhua Xu, \IEEEmembership{Student Member, IEEE}, Yuxiang Sun, \IEEEmembership{Member, IEEE}, and Ming Liu, \IEEEmembership{Senior Member, IEEE} 
\thanks{Manuscript received February 24, 2021; Revised May 23, 2021; Accepted Jun 21, 2021. This paper was recommended for publication by Editor Pauline Pounds upon evaluation of the Associate Editor and Reviewers' comments. 
This work was supported by Collaborative Research Fund by Research Grants Council Hong Kong, under Project No. C4063-18G, Department of Science and Technology of Guangdong Province Fund, under Project No. GDST20EG54 and Zhongshan Municipal Science and Technology Bureau Fund, under project ZSST21EG06, awarded to Prof. Ming Liu. 
\textit{(Corresponding author:  Ming Liu.)}
}
\thanks{Zhenhua Xu is with the Department of Computer Science and Engineering, The Hong Kong University of Science and Technology, Clear Water Bay, Kowloon, Hong Kong (email: zxubg@connect.ust.hk).}
\thanks{Yuxiang Sun is with the Department of Mechanical Engineering, The Hong Kong Polytechnic University, Hung Hom, Kowloon, Hong Kong (email: yx.sun@polyu.edu.hk, sun.yuxiang@outlook.com).}
\thanks{Ming Liu is with the Department of Electronic and Computer Engineering, The Hong Kong University of Science and Technology, Clear Water Bay, Kowloon, Hong Kong (email: eelium@ust.hk).}
\thanks{Digital Object Identifier (DOI): see top of this page.} } 
\IEEEaftertitletext{\vspace{-2\baselineskip}}
  
\begin{document}

\maketitle
\begin{abstract}
Road-boundary detection is important for autonomous driving. It can be used to constrain autonomous vehicles running on road areas to ensure driving safety. Compared with online road-boundary detection using on-vehicle cameras/Lidars, offline detection using aerial images could alleviate the severe occlusion issue. Moreover, the offline detection results can be directly employed to annotate high-definition (HD) maps. In recent years, deep-learning technologies have been used in offline detection. But there still lacks a publicly available dataset for this task, which hinders the research progress in this area. So in this paper, we propose a new benchmark dataset, named \textit{Topo-boundary}, for offline topological road-boundary detection. The dataset contains 25,295 $1000\times1000$-sized 4-channel aerial images. Each image is provided with 8 training labels for different sub-tasks. We also design a new entropy-based metric for connectivity evaluation, which could better handle noises or outliers. We implement and evaluate 3 segmentation-based baselines and 5 graph-based baselines using the dataset. We also propose a new imitation-learning-based baseline which is enhanced from our previous work. The superiority of our enhancement is demonstrated from the comparison.
The dataset and our-implemented code for the baselines are available at \texttt{\url{https://tonyxuqaq.github.io/Topo-boundary/}}.

\vspace{0.25cm}

\end{abstract}

\section{Introduction}
\IEEEPARstart{R}{oad} boundary refers to the dividing line between the road area and off-road area. It can be used to constrain self-driving cars running on road areas, which is important to the safety of autonomous driving.
Currently, most existing works rely on vehicle-mounted sensors (e.g., cameras or Lidars) \cite{lu2020real,zhu2015real,sun20193d,wang2019self} for online road-boundary detection. However, online detection could be severely degraded by occlusions, which is very common in real road environments. Moreover, online detection could be restricted by limited computing resources on vehicles, especially for deep-learning models that require GPU. 
To relieve the aforementioned issues, some works resort to detecting road boundary (or its analogies, such as lane or curb) offline using bird-eye-view (BEV) point-cloud maps or aerial images. As the results are presented in BEV images, they can also be directly utilized for annotating high-definition (HD) maps. So we detect road boundaries offline in this paper. Moreover, our detection results are presented in the form of topological graphs (i.e., vertices and edges). This is important because besides pixel-level annotations, the topological graphs can also identify road-boundary instances and find the spatial connection information of the boundaries.

The published literature working on topological road-boundary detection is very limited. Most of existing works focus on the analogy problem: line-shaped object detection, such as road-lane, road-curb and road-network detection \cite{mattyus2017deeproadmapper,batra2019improved,bastani2018roadtracer,tan2020vecroad,homayounfar2019dagmapper,belli2019image,zhxu2021icurb}. They can be generally divided into two categories: segmentation-based solutions and graph-based solutions. A typical pipeline of the former is first using semantic segmentation to get rough detection results, and then performing heuristic post-processing algorithms on the segmentation maps to refine the results. The latter directly finds the graph structure for the line-shaped objects through iterative graph growing. 
With the great advancement of artificial intelligence, deep-learning technologies are adopted by current existing methods, and great superiority over traditional algorithms is achieved. However, deep-learning-based methods require large-scale datasets to train a model. 
To the best of our knowledge, there is still no publicly available datasets for road-boundary detection in BEV images, which hinders the research process in this area. 
This motivates us to build a large-scale benchmark dataset, named \textit{Topo-boundary}, for road-boundary detection. 

\begin{figure*}[t]
  \centering
    \includegraphics[width=\linewidth]{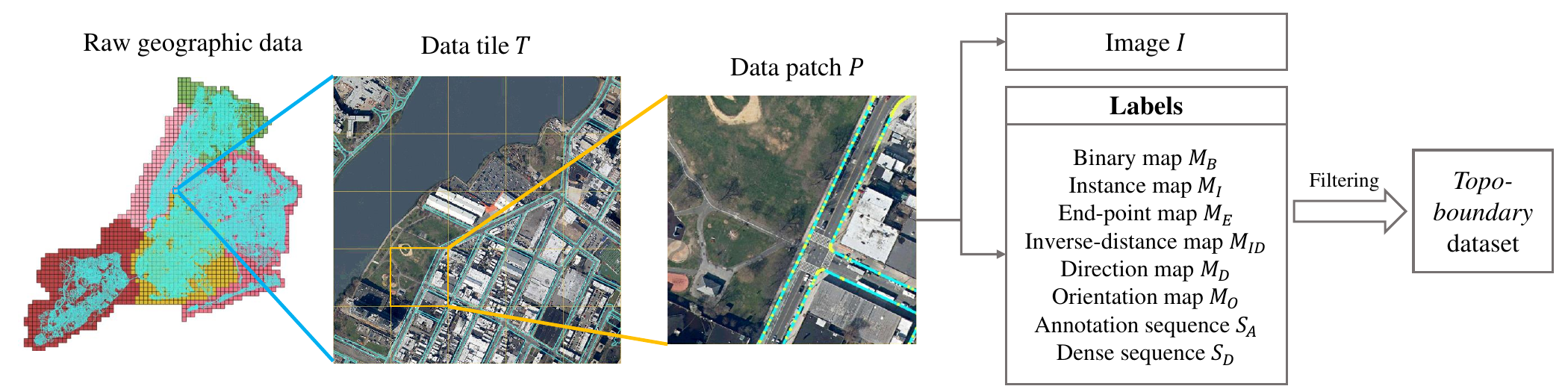}
  \caption{The pipeline to create our dataset from the \textit{NYC Planimetric Database}. The raw geographic data contains 2,147 data tiles. The different colors represent the 5 boroughs and water areas in the New York City. Each tile ($T$) is with $5000\times5000$ size. We split each tile into 25 smaller patches ($T=\{P_i\}^{25}_{i=1}$). Each patch ($P$) contains a 4-channel $1000\times1000$-sized aerial image $I$ and ground-truth road boundaries $G_{gt}$. The $G_{gt}$ in the figure is annotated by polylines, of which edges and vertices are denoted by cyan lines and yellow points, respectively.
  Based on $G_{gt}$, we generate 8 types of ground-truth labels for different tasks. The filtering step is to remove patches based on pre-defined rules. After filtering, 25,295 patches finally remain in our \textit{Topo-boundary}. For better visualization, only RGB channels of the patch image are visualized. Moreover, the width of the ground-truth polylines is increased (the actual width is one pixel). This figure is best viewed in color.}
  \label{pipeline}
\end{figure*}

Our dataset is derived from the \textit{NYC Planimetric Database} \cite{nyc_dataset}.
The raw geographic data (a large whole map, see Fig. \ref{pipeline}) in the \textit{NYC Planimetric Database} covers the whole New York City, including 2,147 $5000\times5000$-sized 4-channel (i.e., red, green, blue and an infrared channel) aerial image tiles (i.e., small squared aerial image constituting the large whole map) and a set of polylines as the road-boundary ground-truth label. 
To build our \textit{Topo-boundary}, we convert the polyline label to graph (i.e., vertices and edges) and split every tile into 25 smaller $1000\times1000$-sized patches. After removing patches with inappropriate annotations, such as patches without road boundaries, 25,295 patches remain. Each patch consists of a 4-channel image and 8 labels for different sub-tasks, such as binary semantic segmentation, orientation learning \cite{batra2019improved}, etc. The dataset can be used for both segmentation-based solutions and graph-based solutions for road-boundary detection. 
 
To facilitate future research on road-boundary detection, we implement and evaluate multiple baselines, including 3 segmentation-based baselines, 5 graph-based baselines. We also design and compare a new imitation-learning-based baseline which is enhanced from our previous work \textit{iCurb} \cite{zhxu2021icurb}. 
For all the baselines, the input is a 4-channel aerial image, the output is the graph representing road boundaries.
Our implemented code for these baselines are open-sourced along with our dataset. 

As line-shaped objects are usually long, thin and irregular, only using pixel-level metrics is not sufficient to evaluate the performance of a method. 
In past works, average path length similarity (APLS) has been widely used to evaluate topology correctness \cite{van2018spacenet}. But since in most cases road boundaries are simple polylines without branches, APLS suffers from randomness and inefficiency. Inspired by the naive connectivity metric proposed in \cite{liang2019convolutional}, we design a new entropy-based connectivity metric (ECM) to evaluate the connectivity of the obtained road-boundary graph, which is effective and more efficient. 
We summarize our contributions here:
\begin{enumerate}
    \item We release the \textit{Topo-boundary} benchmark dataset for topological road-boundary detection using aerial images. To the best of our knowledge, this is the first publicly available dataset for this task.
    \item We propose new evaluation metrics for the task, including relaxed pixel-level metrics and a new entropy-based connectivity metric.
    \item We quantitatively evaluate 9 baseline models using the proposed dataset, which could facilitate comparative study for future methods.
    \item We open-source our-implemented code for the baselines. The code can also be modified with a little effort for other line-shaped object detection.
\end{enumerate}

\section{Related works}
\subsection{Segmentation-based line-shaped object detection}
There are very few segmentation-based works directly detecting road boundaries topologically. So we review several line-shaped object detection works \cite{mattyus2017deeproadmapper,batra2019improved,mnih2010learning}. Volodymyr \textit{et al.} \cite{mnih2010learning} proposed the first deep-learning model with iterative refinement for road-network detection using aerial images. Their solution was further improved by Anil \textit{et al.} \cite{batra2019improved}, in which the authors not only proposed better networks and a training scheme, but also utilized the orientation map to enhance the semantic segmentation results. Different from the above papers using iterative refinement, the solution proposed by Mattyus \textit{et al.} \cite{mattyus2017deeproadmapper} first put forward connection candidates to correct disconnections, and then trained another network to filter candidates. Although segmentation-based solutions are efficient to carry out, they can only produce results at the pixel-level. Moreover, they were often implemented with multi-stage pipelines, so they could not be optimized as a whole, making the graph sensitive to incorrect topology.

\subsection{Graph-based line-shaped object detection}
Taken images as input, graph-based methods can directly output the graph representing target objects. Iterative graph growing is the most commonly used technique, which generates vertices along with the line-shaped object starting from an initial vertex. Past works focusing on other line-shaped objects are similar to our task, such as road-lane detection \cite{homayounfar2018hierarchical,homayounfar2019dagmapper}, road-network detection \cite{bastani2018roadtracer,tan2020vecroad,li2018polymapper} and road-curb detection \cite{zhxu2021icurb}. \textit{RoadTracer} \cite{bastani2018roadtracer} is believed to be the first work in this category. In \cite{bastani2018roadtracer}, the authors trained a multi-layer CNN and successfully extracted the graph of very large road networks. Liang \textit{et al.} \cite{liang2019convolutional} directly worked on road-boundary detection in the BEV point-cloud map. They facilitated semantic segmentation by direction map prediction and proposed cSnake for iterative graph growing. To better solve the graph growing task, line-shaped object detection was first analyzed from the perspective of imitation learning in our previous work \textit{iCurb} \cite{zhxu2021icurb}. \textit{iCurb} presented superiority over other works on road curb detection owing to the training strategy and graph growing policy.

 \begin{figure*}[t]
 \begin{subfigure}[t]{0.135\textwidth}
        \begin{subfigure}[t]{\textwidth}
            \includegraphics[width=\textwidth]{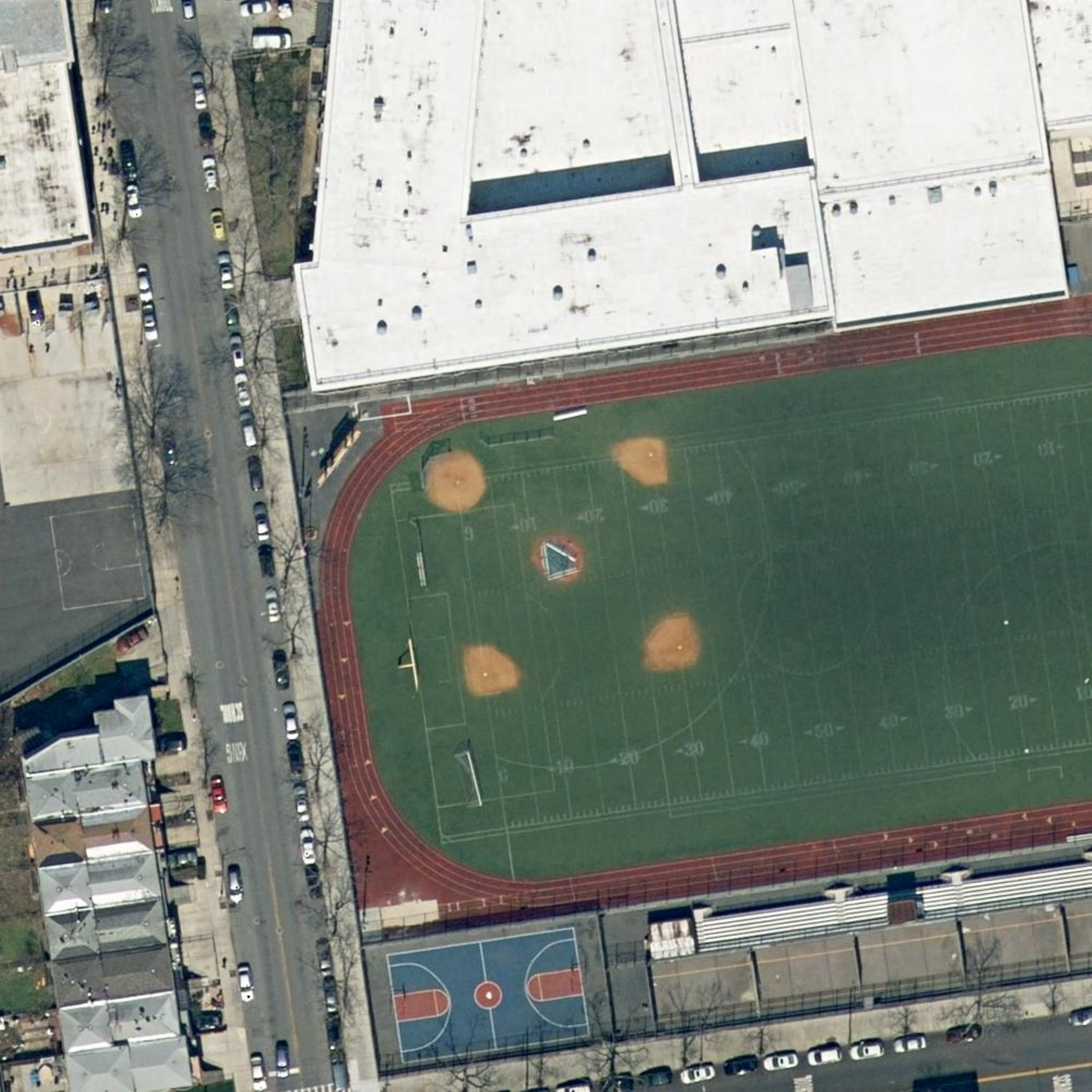}
        \end{subfigure}
        \caption{$I$}
    \end{subfigure}
    \hfill
    \begin{subfigure}[t]{0.135\textwidth}
        \begin{subfigure}[t]{\textwidth}
            \includegraphics[width=\textwidth]{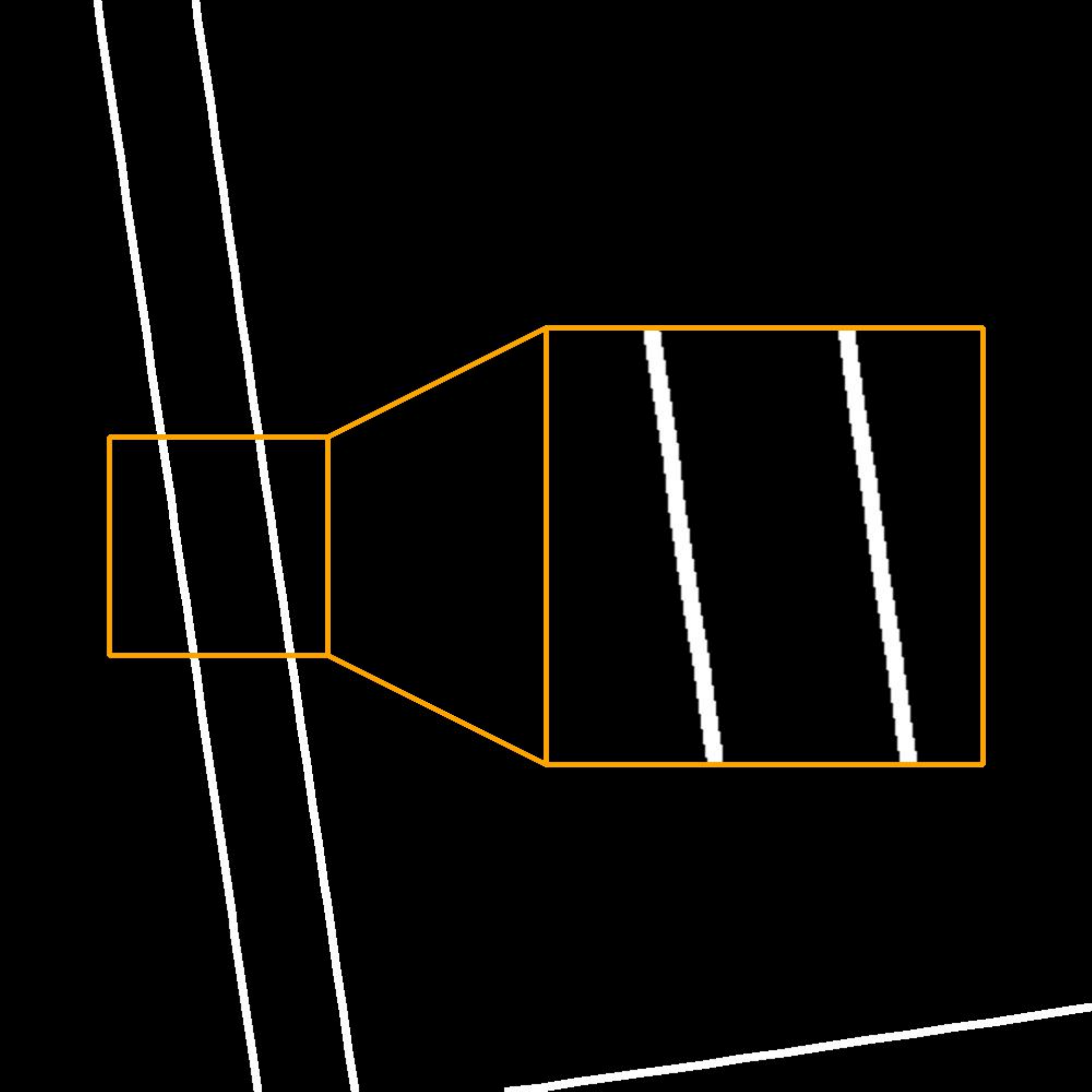}
        \end{subfigure}
        \caption{$M_B$}
    \end{subfigure}
    \hfill
    \begin{subfigure}[t]{0.135\textwidth}
        \begin{subfigure}[t]{\textwidth}
            \includegraphics[width=\textwidth]{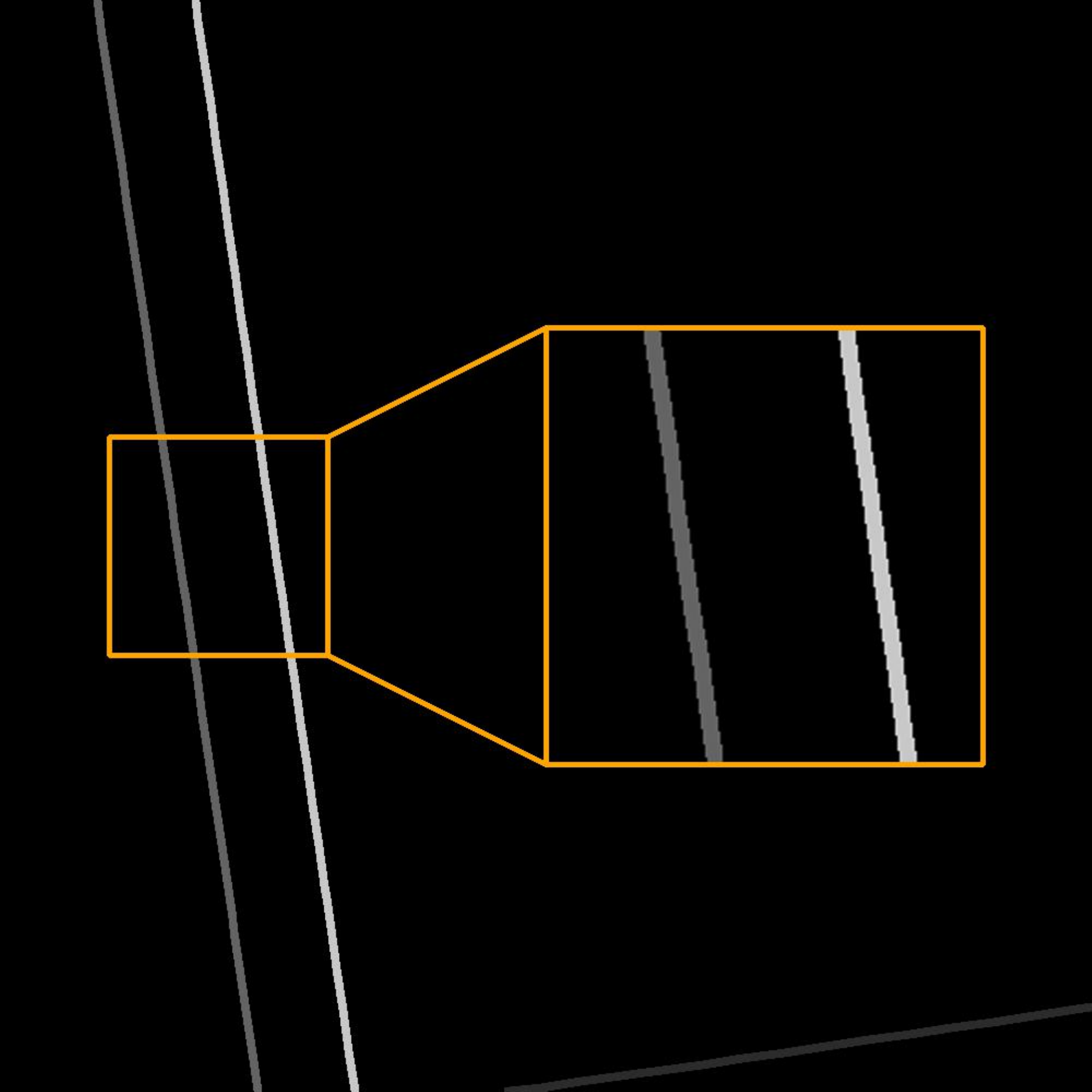}
        \end{subfigure}
        \caption{$M_I$}
    \end{subfigure}
    \hfill
    \begin{subfigure}[t]{0.135\textwidth}
        \begin{subfigure}[t]{\textwidth}
            \includegraphics[width=\textwidth]{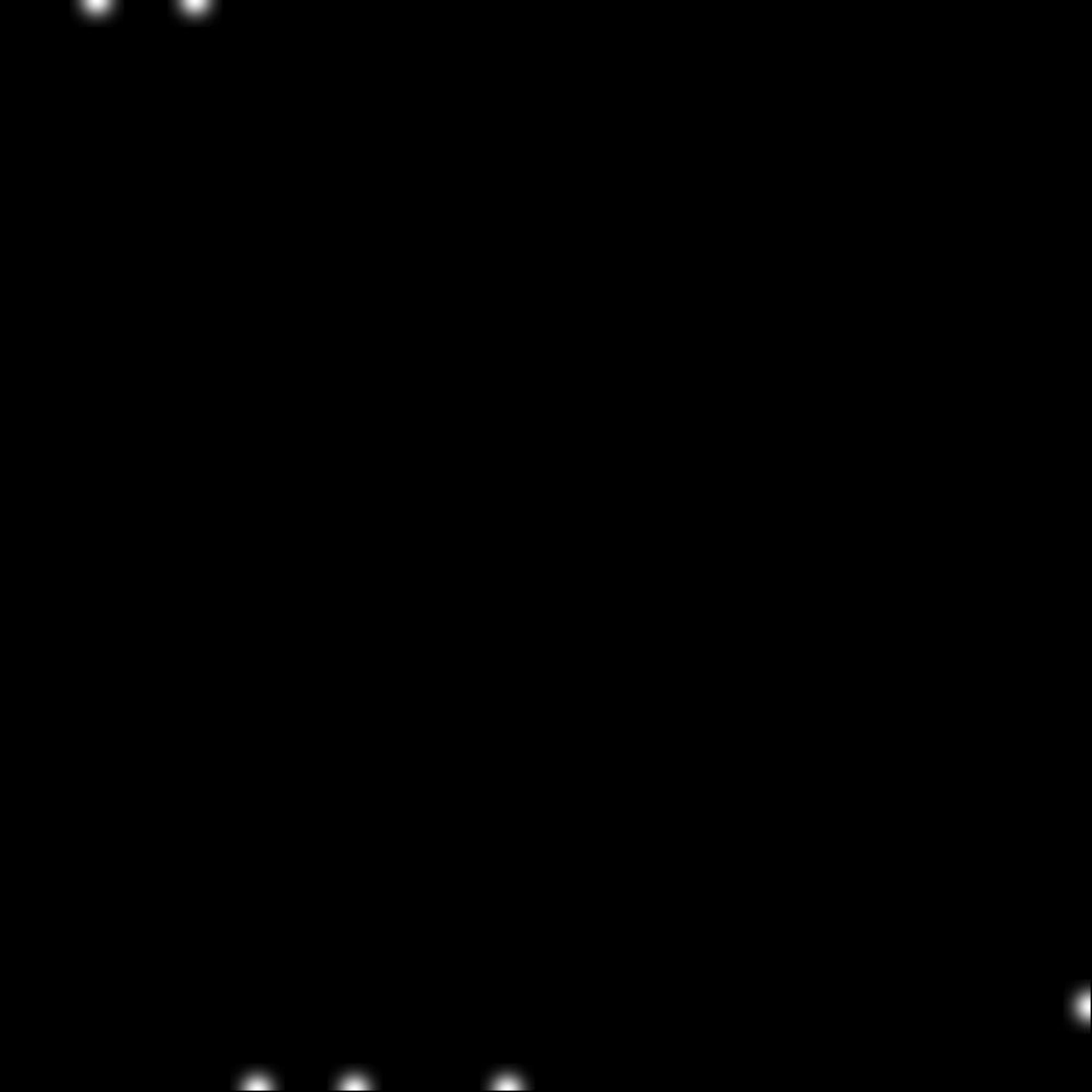}
        \end{subfigure}
        \caption{$M_E$}
    \end{subfigure}
    \hfill
    \begin{subfigure}[t]{0.135\textwidth}
        \begin{subfigure}[t]{\textwidth}
            \includegraphics[width=\textwidth]{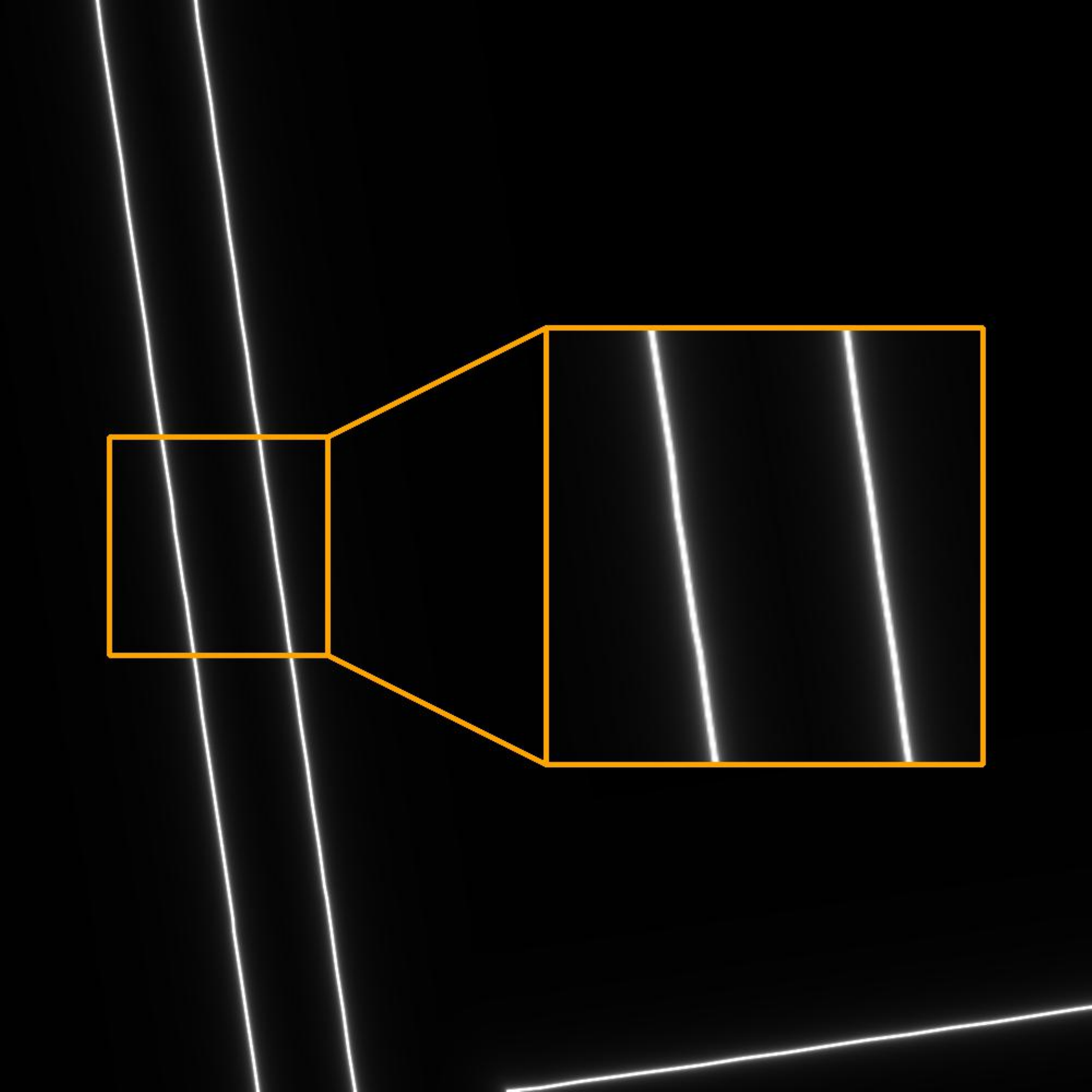}
        \end{subfigure}
        \caption{$M_{ID}$}
    \end{subfigure}
    \hfill
    \begin{subfigure}[t]{0.135\textwidth}
        \begin{subfigure}[t]{\textwidth}
            \includegraphics[width=\textwidth]{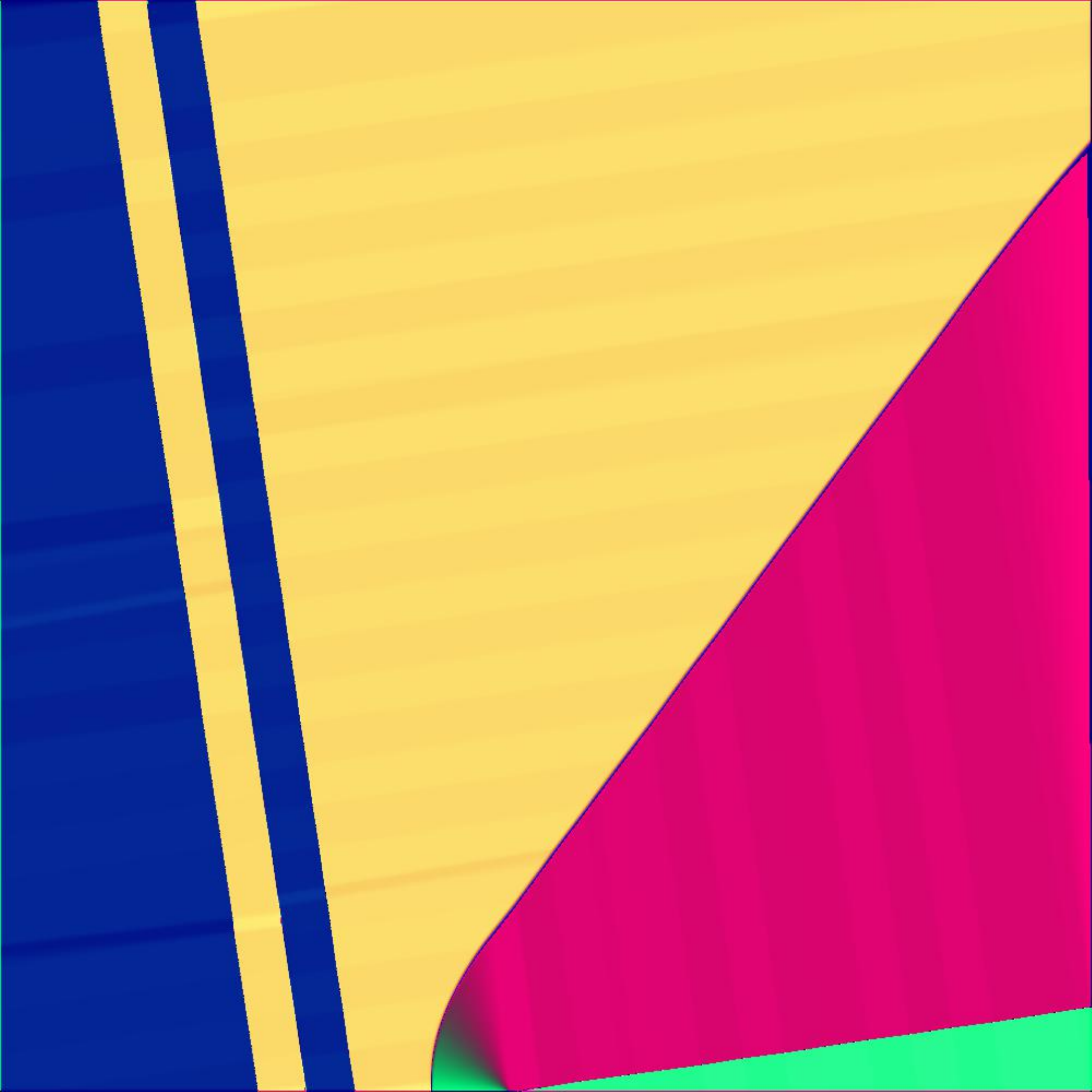}
        \end{subfigure}
        \caption{$M_D$}
    \end{subfigure}
    \hfill
    \begin{subfigure}[t]{0.135\textwidth}
        \begin{subfigure}[t]{\textwidth}
            \includegraphics[width=\textwidth]{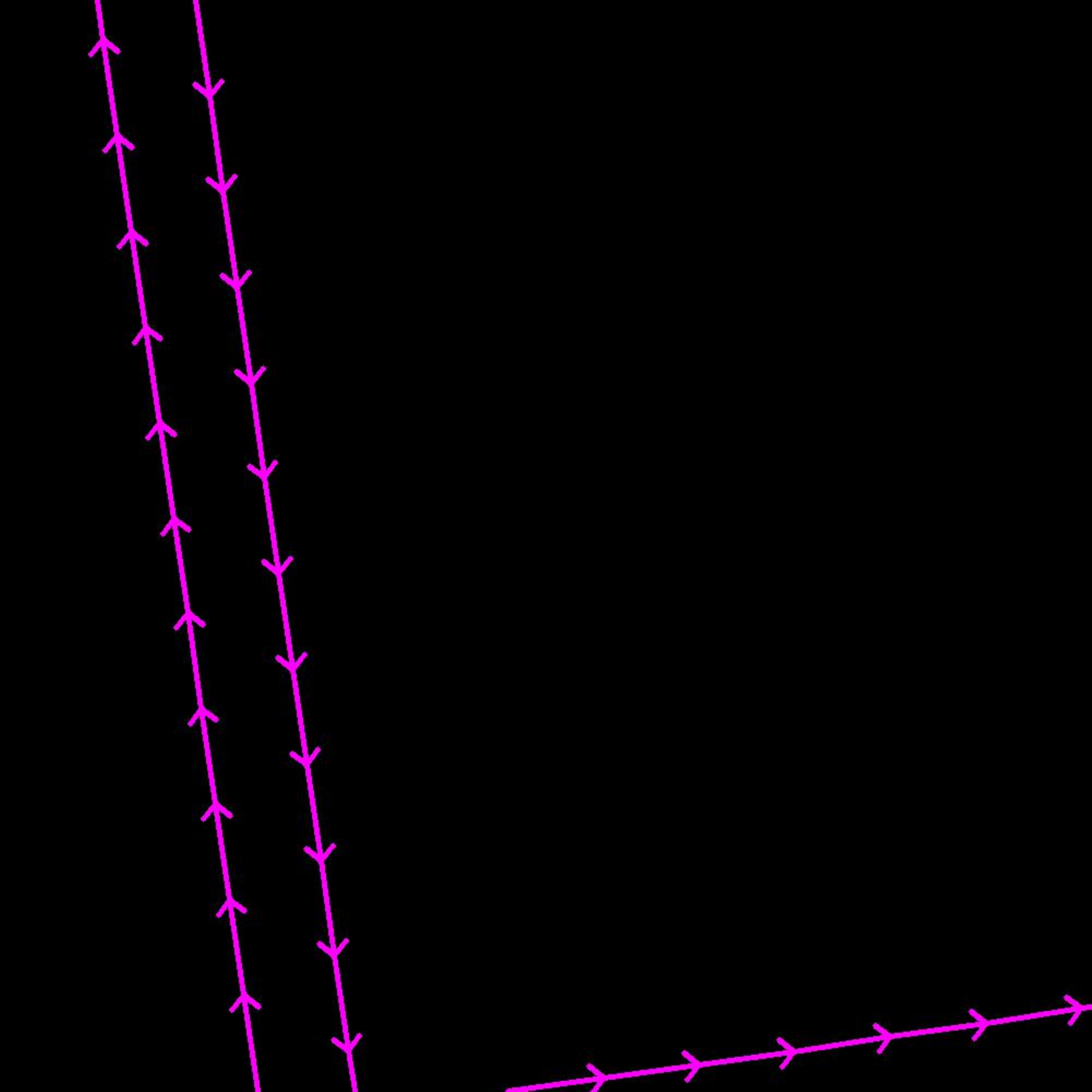}
        \end{subfigure}\vspace{.6ex}
        \caption{$M_O$}
    \end{subfigure}
    \caption{Visualization of label maps for a patch. (a) 4-channel patch image $I$ (only RGB is visualized here); (b) Binary map $M_B$; (c) Instance map $M_I$. Different road boundary instances are labeled with different gray values; (d) End-point map $M_E$; (e) Inverse-distance map $M_{ID}$. Each pixel value is the reciprocal of its shortest distance to the boundary; (f) Direction map $M_{D}$. The red channel is the Sobel derivative of columns, the green channel is the Sobel derivative of rows, and the blue channel is the sum of the other two channels; (g) Orientation map $M_O$. Pink arrows are the schematic demonstration of directional information of the road boundary. For better visualization, the line width in (b) and (c) is increased (the actual width is one pixel).  
    This figure is best viewed in color. Please zoom in for details. For more examples, please see our supplementary document.}
    \label{labels}
\end{figure*}
\subsection{Evaluation metrics for topological correctness }
In the past works on road network detection, topological correctness was usually measured by shortest-path-based metrics, such as APLS \cite{van2018spacenet} and too long/too short (TLTS) similarity \cite{wegner2013higher}. However, due to the random sampling process, these metrics suffer from inefficiency and randomness. Alternatively, in \cite{liang2019convolutional}, a naive metric measuring the connectivity of road boundaries was proposed, but it is too sensitive to noises and may cause incorrect evaluation results. \textit{iCurb} modified the naive metric and addressed the problem to some extent. However, incorrect evaluation still happens. So in this work, we make further revisions to this metric and propose an entropy-based metric for connectivity evaluation.

\subsection{Datasets for line-shaped object detection}
To the best of our knowledge, \cite{liang2019convolutional} is the only published work on road-boundary detection using BEV images. In \cite{liang2019convolutional}, a dataset was created, which contained BEV point-cloud images, RGB images and elevation gradient images. But unfortunately, the dataset was not publicly available. In the road-network detection area, the data was obtained from commercial geospatial datasets \cite{van2018spacenet} or by processing raw geographic images collected from public platforms \cite{haklay2008openstreetmap}. Although these datasets are publicly available, due to the topological differences between road networks and road boundaries, they could not be used in our task.

\section{The Proposed Dataset}
To create our \textit{Topo-boundary} dataset, we generate images and ground-truth labels from the raw geographic data provided by the \textit{NYC Planimetric Database} \cite{nyc_dataset}.
The WGS84 coordinate system is used in \cite{nyc_dataset}, while in our \textit{Topo-boundary}, all the coordinates are transformed to the image coordinate system to facilitate research in this area. Fig. \ref{pipeline} shows the processing pipeline to create our \textit{Topo-boundary}. 
We first split large data tiles into patches, then generate ground-truth labels for each patch, and finally remove inappropriate patches according to predefined rules.

\subsection{Splitting data tiles into patches}
The raw geographic data covers the whole area of the New York City, including 2,147 data tiles. Each tile ($T$) is a $5000\times5000$-sized image with road-boundary ground-truth label. Each pixel represents 1 feet (around 15.2cm). We split each tile into 25 smaller patches ($T=\{P_i\}^{25}_{i=1}$) to reduce the GPU memory cost during training. 
Since our task is mainly performed at the patch level, the subscript $i$ is removed from $P_i$ for expression conciseness in the following text. Each patch ($P$) contains a 4-channel (red, green, blue, infrared, the infrared channel is more sensitive to vegetation and soil) $1000\times1000$-sized aerial image $I$ and ground-truth road boundaries $G_{gt}$. We use $G_{gt}$ to generate 8 labels for different sub-tasks. 
 
 
The raw ground-truth label for road boundaries is the \texttt{PAVEMENT\_EDGE} feature from the \textit{NYC Planimetric Database}. \texttt{PAVEMENT\_EDGE} is manually annotated in the form of polylines (a kind of graph without branches), of which the vertices and edges are recorded using the WGS84 coordinate system. The polylines cover road-surface edges, airport runways and alleys. From the whole map \texttt{PAVEMENT\_EDGE}, we obtain the ground-truth road boundary $G_{gt}$ for each patch $P$.

Unlike tile images, \texttt{PAVEMENT\_EDGE} cannot be directly cut and split because: (1) some edges $e=(v_a,v_b)$ may have one vertex $v_a$ in a patch while the other vertex $v_b$ in another patch; (2) some road boundaries may be cut into multiple pieces by different patches. For the former, we replace the vertex outside the patch (i.e., $v_b$) with the intersection point of $e$ and the patch edge. For the latter, we split the corresponding road boundaries into multiple instances and update \texttt{PAVEMENT\_EDGE}.

After processing \texttt{PAVEMENT\_EDGE}, for each patch, the ground-truth road boundary is still a set of polylines $G_{gt}=\{G_{gt}^i\}^N_{i=1}$. Each polyline represents an road-boundary graph instance $G_{gt}^i=(V_{i},E_{i})$, where $V_{i}$ and $E_{i}$ are the set of vertices and edges in $G_{gt}^i$, respectively. 
Vertices $V_{i}$ are recorded in a list, and then edges in a graph can be easily obtained by connecting two neighboring vertices in the list, thus the edge set $E_i$ is omitted.
Note that $G_{gt}^i$ is sparsely distributed (i.e., adjacent vertices are not eight-neighboring to each other in the image coordinate system). 

\subsection{Generating ground-truth labels for each patch}
For each patch, we generate 8 different types of labels for different tasks. The labels are: annotation sequence $S_A$, dense sequence $S_D$, binary map $M_B$, instance map $M_I$, endpoint map $M_E$, inverse-distance map $M_{ID}$, direction map $M_{D}$ and orientation map $M_O$. All the label maps are generated from ground-truth road boundaries ($G_{gt}$) within the current patch by our processing algorithms, and are recorded in the image coordinate system of the current patch.

\subsubsection{Annotation sequence and dense sequence}
\begin{figure}[t]
  \centering
    \includegraphics[width=\linewidth]{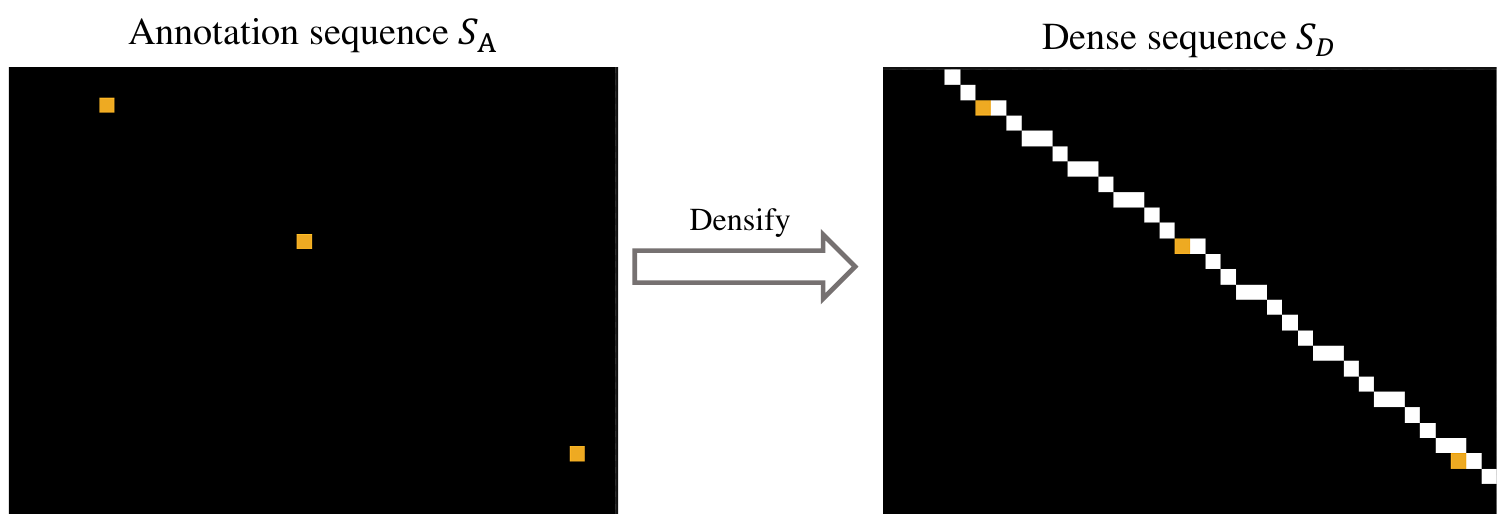}
  \caption{The sequence densification process for a part of a sample patch. The orange pixels represent vertices in the annotation sequence $S_A$. They are interpolated to realize every two adjacent vertices eight-neighboring to each other. The interpolation result is the dense sequence $S_D$. $S_A$ can be regarded as a subset of $S_D$, which contains only the key vertices. The figure is best viewed in color.}
  \label{densify}
\end{figure}

As the ground-truth road boundaries are annotated vertex-by-vertex as chain sequences, the vertices of a road boundary instance $G_{gt}^i$ are ordered starting from an initial vertex to the end vertex. We call the ordered vertices of $G_{gt}^i$ as the annotation sequence $S_A$. $S_A$ is critical to generate other labels and can be applied to train graph-based solutions. However, graph-based solutions trained by $S_A$ suffer from the \textit{teacher forcing} problem \cite{ranzato2015sequence}, which is caused by the data distribution mismatch between the training and inference period. It is also analyzed in \cite{bastani2018roadtracer}. To address this problem, the training label for iterative graph generation should be generated on-the-fly. Therefore, $S_A$ needs to be densified. For every two adjacent vertices in $S_A$, interpolation is conducted to obtain the dense sequence $S_D$. In $S_D$, any two adjacent vertices are eight-neighboring to each other. The densification process is illustrated in Fig. \ref{densify}.
 
\subsubsection{Binary map, instance map and endpoint map}
Binary map $M_B$ is commonly used for semantic segmentation. The dense sequence $S_D$ can be directly used to generate $M_B$ by setting all the pixels covered by $S_D$ as 1 (i.e., foreground pixels) and the other pixels as 0 (i.e., background pixels). In this way, we can get the binary map of a patch and its topological correctness can be guaranteed. The instance map $M_I$ can be obtained similarly, but pixels covered by different instances are multiplied with instance IDs. 

To facilitate segmentation and generate candidate initial vertices for iterative graph generation, endpoint prediction is required in some solutions \cite{liang2019convolutional,zhxu2021icurb}. For each road boundary instance $G_{gt}^i$ in a patch, there are 2 endpoints. We multiply each endpoint by a Gaussian kernel function.
 
\subsubsection{Inverse-distance map and direction map}
Inverse-distance map ($M_{ID}$) prediction is first proposed in \cite{liang2018end} to facilitate semantic segmentation. The value of each pixel in this map is the reciprocal of its shortest distance to the ground-truth road boundary. Compared with the binary map $M_B$, $M_{ID}$ is more informative considering that all the pixels of $M_{ID}$ are encoded with information. The generation of $M_{ID}$ is accelerated by GPU parallelism in our implementation.
 
To further make use of the spatial information in $M_{ID}$, the authors in \cite{liang2019convolutional} generate the direction map $M_D$ based on $M_{ID}$. The direction map is a vector field ($M_D\in \mathbb{R}^{2\times H\times W}$) of normal directions to the road boundary. It is calculated by taking the Sobel derivative of $M_{ID}$ followed by a normalization step. Intuitively, $M_D$ records the direction to the closest road boundary of each pixel by unit vectors. The application of $M_D$ and $M_{ID}$ encourages the neural network to focus on road boundaries. 
 \begin{figure}[t]
 \begin{subfigure}[t]{0.155\textwidth}
        \begin{subfigure}[t]{\textwidth}
            \includegraphics[width=\textwidth]{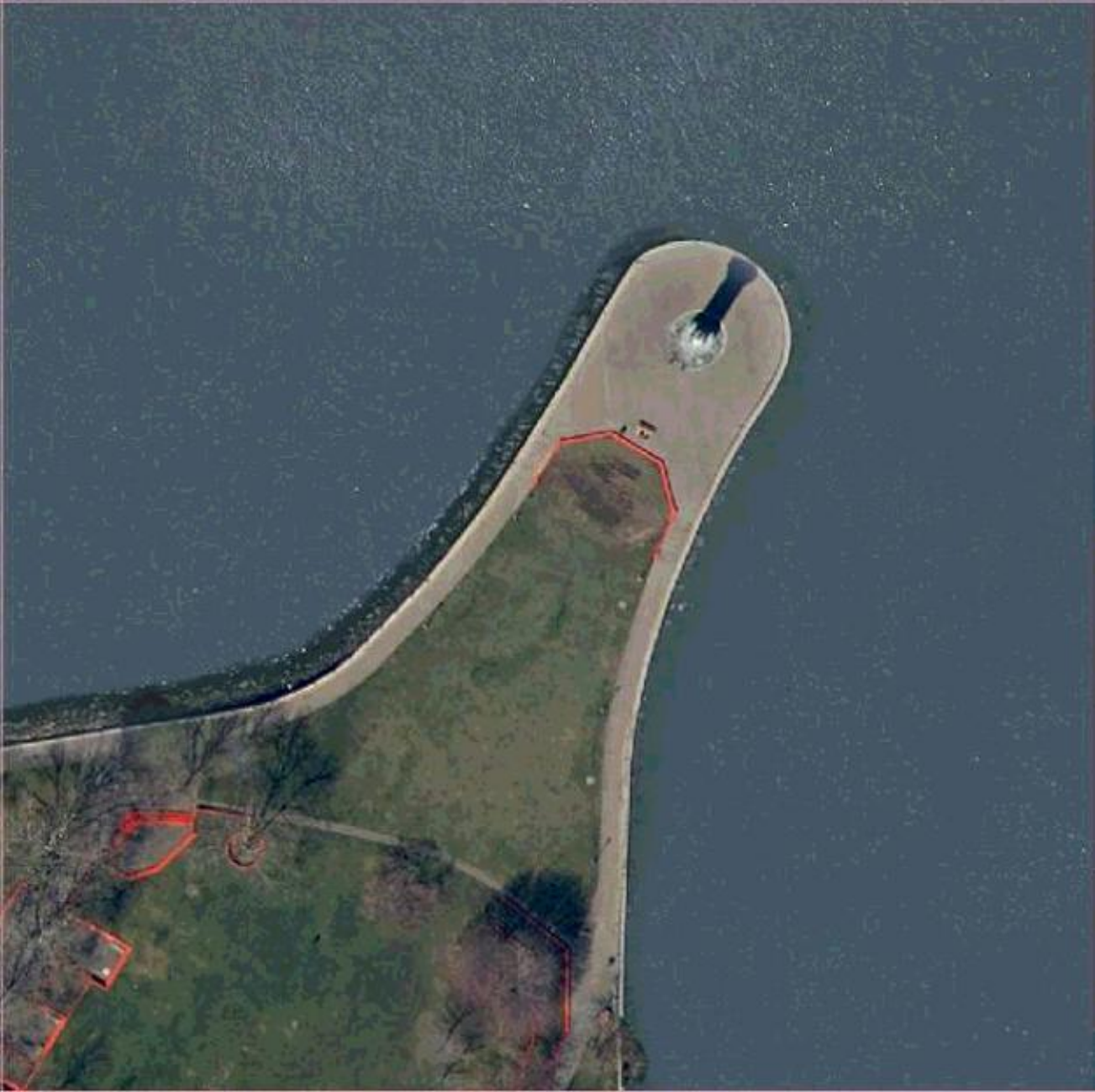}
        \end{subfigure}
        \caption{}
    \end{subfigure}
    \hfill
    \begin{subfigure}[t]{0.155\textwidth}
        \begin{subfigure}[t]{\textwidth}
            \includegraphics[width=\textwidth]{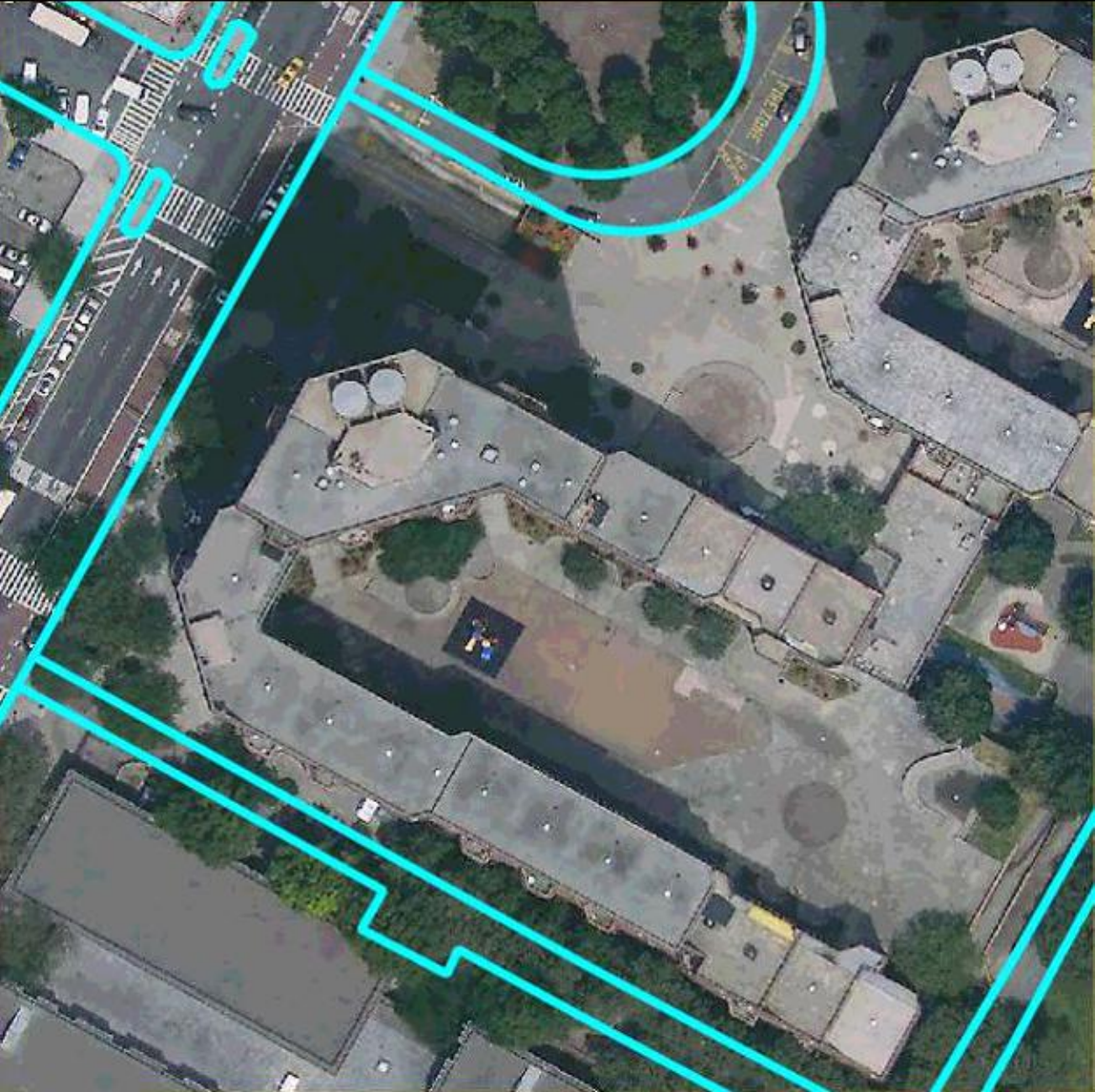}
        \end{subfigure}
        \caption{}
    \end{subfigure}
    \hfill
    \begin{subfigure}[t]{0.155\textwidth}
        \begin{subfigure}[t]{\textwidth}
            \includegraphics[width=\textwidth]{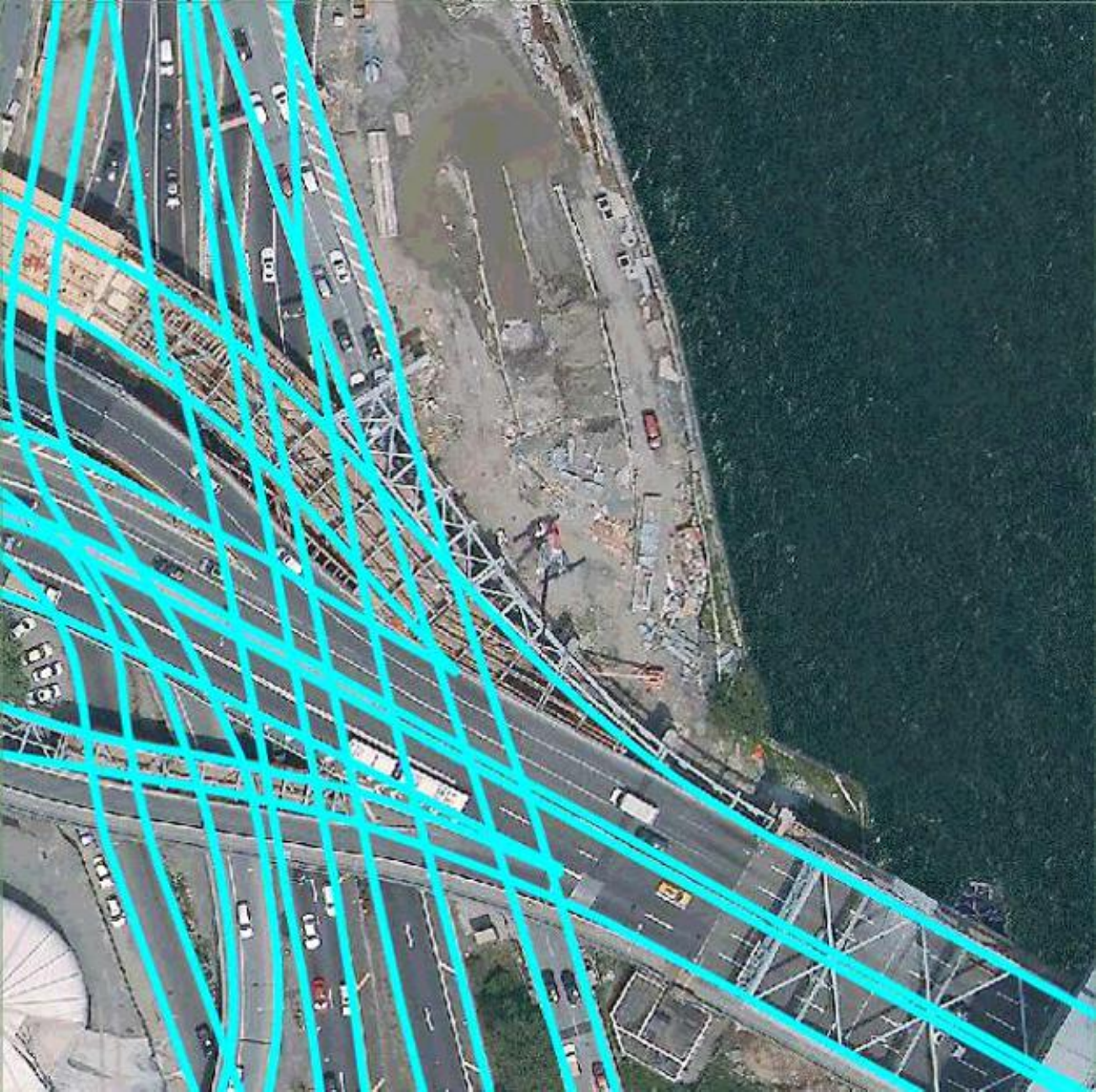}
        \end{subfigure}
        \caption{}
    \end{subfigure}
    \caption{Sample inappropriate patches. The cyan lines represent the ground-truth polylines for road boundaries. (a) A patch without road boundaries inside. Such case happens frequently near ocean and suburb areas; (b) A patch in which ground-truth road boundaries have intersection points. This is usually caused by alleys that snapped to the road boundary. Intersection points are not reasonable considering the topological characteristics of the road boundary; 
    (c) A patch with very complicated scenarios. For these patches, few methods can obtain reasonable results, especially for graph-based methods. Therefore they are removed at this stage. 
    For better visualization, only RGB channels of the patch image are visualized, and the width of the ground-truth polylines is increased (the actual width is one pixel).}
    \label{filter}
\end{figure}

\subsubsection{Orientation map}
Orientation learning \cite{batra2019improved} makes use of the direction and connection information of road networks, which greatly enhances the performance of semantic segmentation. Thus we provide the orientation map for this task to assist road-boundary detection. Due to the bi-direction nature of road boundaries, we regard them as undirected graphs. For each road boundary instance $G_{gt}^i$, we randomly select one end vertex as the starting point while the other as the ending point. Then from the starting point, for every two adjacent vertices, we calculate a directional vector and convert it to a radian value $r$. For all pixels covered by the edge connecting these two adjacent vertices, their values are set to $r$. Intuitively, the value of each foreground pixel is the radian value of the edge it belongs to. Fig. \ref{labels} displays all the label maps for a sample patch.

\subsection{Removing patches according to pre-defined rules} 
We remove the inappropriate patches according to the following predefined rules:
\begin{enumerate}
    \item The patch with no road boundary inside. Such a case is quite common, especially in suburb area images;
    \item The patch with intersection points. In the raw ground-truth label of the road boundary, some alleys are snapped to it and cause intersections, which is not appropriate. Thus we remove these patches;
    \item The patch that has very complex scenarios, such as overlapping interchanges. 
\end{enumerate}

The cases in the removed patches would happen in city-scale real-world applications. So, this could be treated as a limitation of our work at this stage. Fig. \ref{filter} shows sample inappropriate patches.

\subsection{Data splitting}
After filtering, there are finally 25,295 patches remaining in our \textit{Topo-boundary}. 
We randomly split these patches into a training set (10,236 patches), a validation set (1,770 patches), a testing set (3,289 patches) and a pretraining set (10,000 patches). The pretraining set is used for multi-stage methods, like \cite{mattyus2017deeproadmapper}, or to pretrain the feature extraction module of the graph-based solutions to accelerate the training convergence, such as \cite{homayounfar2019dagmapper}. If pretraining is not needed, the samples in this set could be used for training. The boroughs of New York City (e.g., Queens or Manhattan) is not considered during data splitting. The above data splitting scheme is adopted in the experiments of this paper. Users are free to split the dataset according to their needs.

\section{Evaluation metrics}
To ensure comprehensive and fair comparison, we design our evaluation metrics, including 3 relaxed pixel-level metrics (i.e., Precision, Recall and F1-score), the naive connectivity metric \cite{liang2019convolutional}, APLS \cite{van2018spacenet} and an entropy-based connectivity metric (ECM) for topological correctness measurement. For all the metrics, larger values indicate better performance.

\subsection{Pixel-level metrics}
Pixel-level metrics (i.e., Precision, Recall and F1-score) measure the prediction accuracy of every pixel. Unlike most past works that directly compare the prediction results with the ground truth, in this work we use the relaxed version of these metrics following \cite{homayounfar2019dagmapper,liang2019convolutional,zhxu2021icurb}.

Let $G_{pre}$ denote the predicted graph and $G_{gt}$ denote the ground-truth graph. Note that both graphs have been densified. Precision is the ratio of pixels in $G_{pre}$ that fall within $\tau$ distance to $G_{gt}$. Recall is the ratio of pixels in $G_{gt}$ that fall within $\tau$ distance to $G_{pre}$.
Distance $\tau$ reflects the tolerance of inaccuracy. If $\tau=1$, the relaxed pixel-level metrics degenerate into the commonly-used hard pixel-level metrics. In this paper, we report the evaluation results with $\tau$ as 2, 5 and 10 pixels, respectively.

\subsection{The proposed entropy-based connectivity metric (ECM)}
    \begin{figure}[t]
 \begin{subfigure}[t]{0.155\textwidth}
        \begin{subfigure}[t]{\textwidth}
            \includegraphics[width=\textwidth]{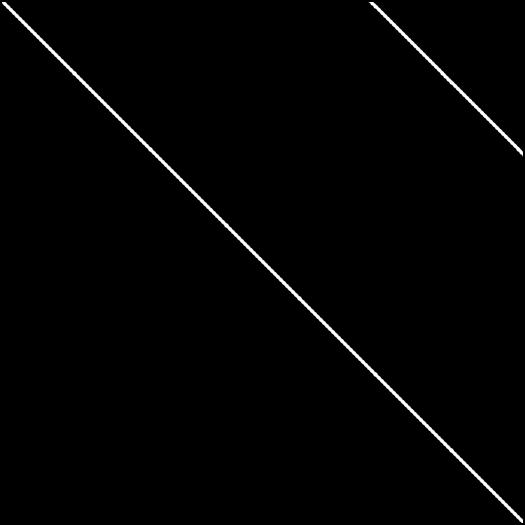}
        \end{subfigure}
        \caption{Ground truth}
    \end{subfigure}
    \begin{subfigure}[t]{0.155\textwidth}
        \begin{subfigure}[t]{\textwidth}
            \includegraphics[width=\textwidth]{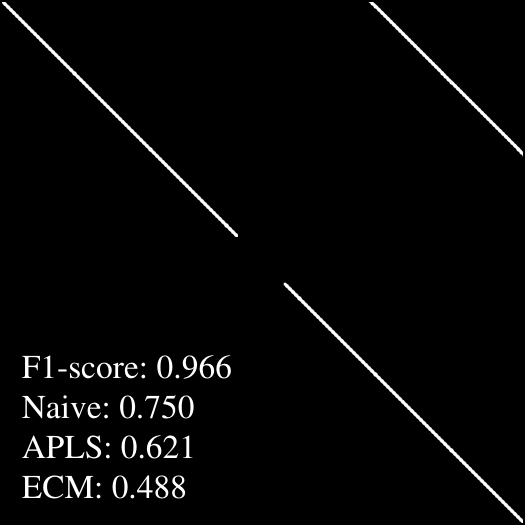}
        \end{subfigure}
        \caption{Prediction 1}
    \end{subfigure}
    \begin{subfigure}[t]{0.155\textwidth}
        \begin{subfigure}[t]{\textwidth}
            \includegraphics[width=\textwidth]{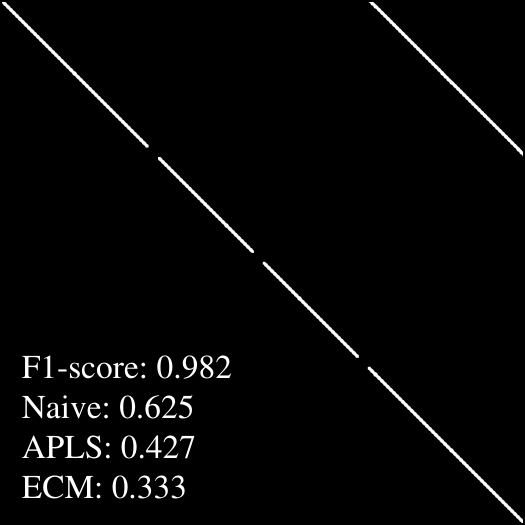}
        \end{subfigure}
        \caption{Prediction 2}
    \end{subfigure}
    \vfill
   \begin{subfigure}[t]{0.155\textwidth}
        \begin{subfigure}[t]{\textwidth}
            \includegraphics[width=\textwidth]{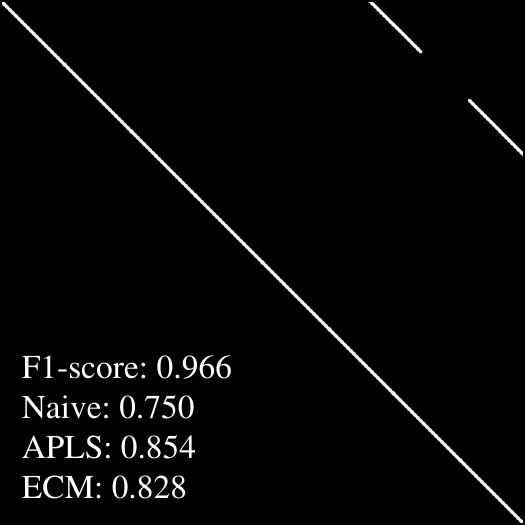}
        \end{subfigure}
        \caption{Ground truth 3}
    \end{subfigure}
    \begin{subfigure}[t]{0.155\textwidth}
        \begin{subfigure}[t]{\textwidth}
            \includegraphics[width=\textwidth]{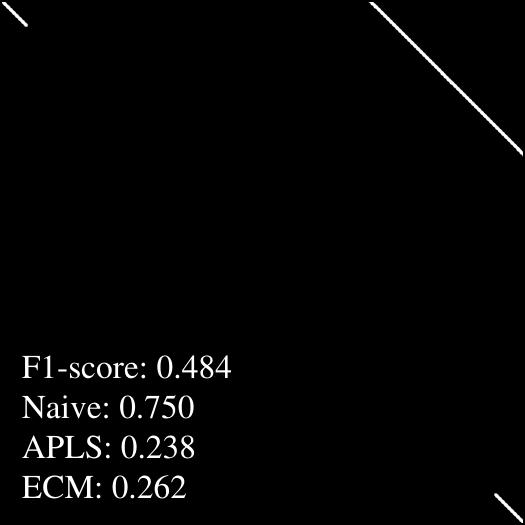}
        \end{subfigure}
        \caption{Prediction 4}
    \end{subfigure}
    \begin{subfigure}[t]{0.155\textwidth}
        \begin{subfigure}[t]{\textwidth}
            \includegraphics[width=\textwidth]{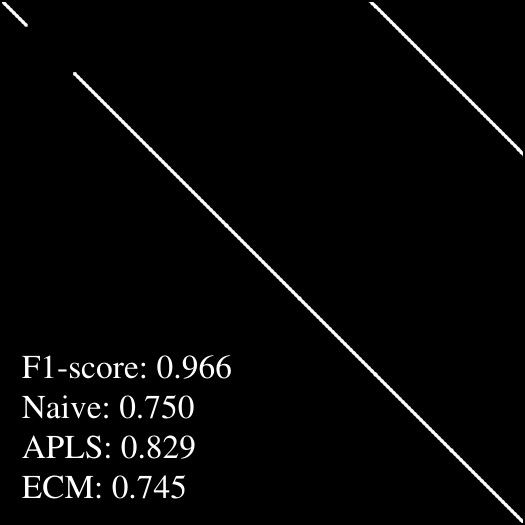}
        \end{subfigure}
        \caption{Prediction 5}
    \end{subfigure}
    \caption{Comparison of topology metrics. (a) The ground truth; (b)-(f) are predictions with different incorrect disconnections. The evaluation score of each metric is shown in prediction sub-figures. (b)(c) reflects whether metrics suffice principle 1: (c) has more disconnections, thus the connectivity score of (b) should be larger than that of (c). All three metrics meet the requirement. Similarly, (b)(d), (b)(e), (b)(f) reflect whether metrics suffice principle 2,3 and 4, respectively. From the comparison, we find that both APLS and ECM work well for all cases, while the naive connectivity metric and pixel-level metrics fail to follow the principles. Please zoom in for details.}
    \label{entropy}
\end{figure}
Assume the predicted road boundary instances are $G_{pre}=\{G_{pre}^j\}_{j=1}^M$ and the ground-truth instances are $G_{gt}=\{G_{gt}^i\}_{i=1}^N$. In our task, topology errors mainly refer to disconnections. Therefore, the topology correctness metric should suffice the following principles: (1) Punish incorrect disconnections; (2) Assign shorter ground-truth instances $G_{gt}^i$ less weights; (3) Longer incorrect disconnection receives a larger penalty; (4) The prediction whose dominant $G_{pre}^j$ has higher dominance should receive a higher score. Dominant predicted instance $G_{pre}^j$ is the instance with the highest dominance value, where \textit{dominance value} is the ratio of the length of $G_{pre}^j$ to the sum of the length of predicted instances assigned to $G_{gt}^i$. 

Among metrics used in the past work, the widely applied path-based metrics, like APLS and TLTS, meet almost all aforementioned requirements. However, they suffer from randomness and inefficiency due to the random sampling process. Because of this, in \cite{liang2019convolutional}, an alternative naive connectivity metric is proposed. The authors first match instances in $G_{pre}$ with instances in $G_{gt}$ by minimizing the Hausdorff distance \cite{huttenlocher1993comparing}. If an instance $G_{pre}^j$ matches with $G_{gt}^i$, $G_{pre}^j$ is assigned to $G_{gt}^i$. For each ground-truth instance $G_{gt}^i$, let $M_i$ denote the number of assigned predicted instances. Then the connectivity of instance $G_{gt}^i$ is $C_i=\frac{1(M_i>0)}{M_i}$, and the final connectivity of the whole patch is the average sum of $C_i$ of all ground-truth instances.

This naive metric can reflect the connectivity of the prediction to some extent, but it fails to meet most aforementioned principles. To relieve this problem, in our previous work \textit{iCurb} \cite{zhxu2021icurb}, we made some adjustments to the naive metric. The matching method is changed to a voting scheme. Each pixel of $G_{pre}^j$ finds its Euclidean-closest $G_{gt}^i$ and makes a vote, then $G_{pre}^j$ is assigned to the ground-truth instance $G_{gt}^i$ that wins the most votes. This voting scheme is more stable than the Hausdorff distance that is sensitive to noises. In addition, a weighting hyper-parameter that gives the longer $G_{gt}^i$ larger weights is assigned to each $G_{gt}^i$. However, the connectivity metric from the \textit{iCurb} paper cannot meet all the requirements.

Therefore, in this paper, we further modify the connectivity metric and propose an entropy-based connectivity metric (ECM), which is shown in the following equation:
\begin{equation}
\begin{aligned}
    ECM &= \sum_{i=1}^N \alpha_i e^{-C_i}, \ \ \text{where} \  
    C_i = \sum_{j=1}^{M_i}-p_jlog(p_j),
\end{aligned}
\end{equation}
where $C_i$ is the connectivity of the $i-th$ ground-truth instance $G_{gt}^i$; $\alpha_i$ is the completion of $G_{gt}^i$, which is equal to the sum of the length of assigned instances in $G_{pre}$ projected onto $G_{gt}^i$ divided by the length of $G_{gt}^i$; $N$ is the total number of $G_{gt}^i$ in the current patch; $M_i$ is the number of predicted road boundary instances $G_{pre}^j$ that are assigned to $G_{gt}^i$, and $p_j$ is the dominance of $G_{pre}^j$. Entropy is a definition created by Shannon in information theory to measure information content. An event with greater uncertainty tends to have larger entropy. If the dominant predicted instance $G_{pre}^{j0}$ has large dominance value, $G_{gt}^i$ should have good connectivity measure $C_i$ since $G_{pre}^{j0}$ can well approximate $G_{gt}^{i}$ with more certainty. In 
summary, ECM assigns longer predicted graph instances with larger weights to calculate $C_i$, so that short instance could not greatly affect the final evaluation results. Therefore, ECM shows better ability to handle noises and outliers. 
An example comparing different connectivity metrics is shown in Fig. \ref{entropy}.

\section{Baseline models}
In this work, we implement several segmentation-based baselines and graph-based baselines. We also design a new imitation-learning-based method which is enhanced from our previous work \textit{iCurb} \cite{zhxu2021icurb}.

\subsection{Segmentation-based baselines}
 
\subsubsection{Naive baseline} This baseline is proposed by ourselves. Firstly, we employ U-net \cite{ronneberger2015u} to obtain the segmentation map of road boundaries. Then, the segmentation results are refined by filtering noisy segments. Finally, the refined segmentation map is skeletonized to get the graph for the road boundaries.
    
\subsubsection{DeepRoadMapper} This baseline (ICCV2017) \cite{mattyus2017deeproadmapper} proposes to fix disconnections in road network predictions. Due to the similarity between our task and road network detection, we adapt their method for road-boundary detection.
    
\subsubsection{OrientationRefine} This baseline (CVPR2019) \cite{batra2019improved} utilizes the orientation map to enhance the segmentation of road networks and achieves improvements. Besides, an extra network is trained to refine the segmentation results iteratively. This work can be directly used for our task without many modifications.


\subsection{Graph-based baselines}  
\subsubsection{RoadTracer} This baseline (CVPR2018) \cite{bastani2018roadtracer} is believed to be the first work that solves the line-shaped object detection problem by an iterative graph generation approach. However, the network for feature extraction is merely a multi-layer CNN without skip connections. 
    
\subsubsection{VecRoad} This baseline (CVPR2020) \cite{tan2020vecroad} greatly improves \textit{RoadTracer}, but follows the same core idea. Res2Net \cite{gao2019res2net} is utilized as the backbone for multi-scale feature extraction. However, there is a limited improvement to the training strategy. 
Both \textit{VecRoad} and \textit{RoadTracer} work on road network detection, which has different topological characteristics with the road boundary (i.e., road boundaries are scattered polylines without branches while the road network is a connected graph with many intersections), thus we modified their exploration algorithm and the overall pipeline to make them applicable for our task. Besides, road network detection only requires coarse detection of the road centerline, while road boundary detection expects the fine structure of both sides of the road. Therefore, our task has higher demands compared with road network detection.
    
\subsubsection{ConvBoundary} This baseline (CVPR2019) \cite{liang2019convolutional} is the only graph-based work that directly focuses on road-boundary detection. It is a two-stage solution. First, it predicts the inverse-distance map, endpoint map and direction map simultaneously. Then, based on these three maps, a model named cSnake is trained to generate the final graph vertex-by-vertex.
    
\subsubsection{DagMapper} This baseline (ICCV2019) \cite{homayounfar2019dagmapper} is designed for road-lane detection. The input of both \textit{DagMapper} and \textit{ConvBoundary} are BEV images of pre-built point-cloud map. Compared with aerial images, BEV images of point-cloud map has much higher resolution and simpler scenarios. For example, the occlusion issue in the point-cloud map is relieved and only the near-road area is covered, which could be regarded as a kind of attention. However, the point-cloud map is time-consuming and expensive to obtain and update, while aerial images are more and more publicly available all over the world. Therefore, our task is more suitable for wide-area applications.
    
\subsubsection{iCurb} This baseline (RAL2021) \cite{zhxu2021icurb} is the first work that solves the iterative graph generation from the perspective of imitation learning. \textit{iCurb} has a DAgger-based \cite{ross2011reduction} training strategy, which greatly enhances the performance. This work focuses on road-curb detection, which is a subclass of road boundary, thus it can be directly applied to our task.  

Except \textit{RoadTracer}, the code of all the graph-based baselines are not publicly available. So we implement them by ourselves.
 \begin{figure*}[t]
    \newcommand{\picvisi}[1]{000167#141}
    \newcommand{\picvisii}[1]{000197#122}
    \newcommand{\picvisiii}[1]{000230#121}
    \newcommand{\picvisiv}[1]{000217#143}
 \centering
    \begin{subfigure}[t]{0.093\textwidth}
        \begin{subfigure}[t]{\textwidth}
            \includegraphics[width=\textwidth]{img/gt_\picvisi{_}.pdf}
        \end{subfigure}\vspace{.6ex}
        \begin{subfigure}[t]{\textwidth}
            \includegraphics[width=\textwidth]{img/gt_\picvisii{_}.pdf}
        \end{subfigure}\vspace{.6ex}
        \begin{subfigure}[t]{\textwidth}
            \includegraphics[width=\textwidth]{img/gt_\picvisiii{_}.pdf}
        \end{subfigure}\vspace{.6ex}
        \begin{subfigure}[t]{\textwidth}
            \includegraphics[width=\textwidth]{img/gt_\picvisiv{_}.pdf}
        \end{subfigure}\vspace{.6ex}
        \caption{GT}
        \label{fig_qualitative_1st}
    \end{subfigure}
    \begin{subfigure}[t]{0.093\textwidth}
        \begin{subfigure}[t]{\textwidth}
            \includegraphics[width=\textwidth]{img/naive_\picvisi{_}.pdf}
        \end{subfigure}\vspace{.6ex}
        \begin{subfigure}[t]{\textwidth}
            \includegraphics[width=\textwidth]{img/naive_\picvisii{_}.pdf}
        \end{subfigure}\vspace{.6ex}
        \begin{subfigure}[t]{\textwidth}
            \includegraphics[width=\textwidth]{img/naive_\picvisiii{_}.pdf}
        \end{subfigure}\vspace{.6ex}
        \begin{subfigure}[t]{\textwidth}
            \includegraphics[width=\textwidth]{img/naive_\picvisiv{_}.pdf}
        \end{subfigure}\vspace{.6ex}
        \caption{Naive}
        \label{fig_qualitative_1st}
    \end{subfigure}
    \begin{subfigure}[t]{0.093\textwidth}
        \begin{subfigure}[t]{\textwidth}
            \includegraphics[width=\textwidth]{img/dr_\picvisi{_}.pdf}
        \end{subfigure}\vspace{.6ex}
        \begin{subfigure}[t]{\textwidth}
            \includegraphics[width=\textwidth]{img/dr_\picvisii{_}.pdf}
        \end{subfigure}\vspace{.6ex}
        \begin{subfigure}[t]{\textwidth}
            \includegraphics[width=\textwidth]{img/dr_\picvisiii{_}.pdf}
        \end{subfigure}\vspace{.6ex}
        \begin{subfigure}[t]{\textwidth}
            \includegraphics[width=\textwidth]{img/dr_\picvisiv{_}.pdf}
        \end{subfigure}\vspace{.6ex}
        \caption{\cite{mattyus2017deeproadmapper}}
        \label{fig_qualitative_1st}
    \end{subfigure}
    \begin{subfigure}[t]{0.093\textwidth}
        \begin{subfigure}[t]{\textwidth}
            \includegraphics[width=\textwidth]{img/ic_\picvisi{_}.pdf}
        \end{subfigure}\vspace{.6ex}
        \begin{subfigure}[t]{\textwidth}
            \includegraphics[width=\textwidth]{img/ic_\picvisii{_}.pdf}
        \end{subfigure}\vspace{.6ex}
        \begin{subfigure}[t]{\textwidth}
            \includegraphics[width=\textwidth]{img/ic_\picvisiii{_}.pdf}
        \end{subfigure}\vspace{.6ex}
        \begin{subfigure}[t]{\textwidth}
            \includegraphics[width=\textwidth]{img/ic_\picvisiv{_}.pdf}
        \end{subfigure}\vspace{.6ex}
        \caption{\cite{batra2019improved}}
        \label{fig_qualitative_1st}
    \end{subfigure}
    \begin{subfigure}[t]{0.093\textwidth}
        \begin{subfigure}[t]{\textwidth}
            \includegraphics[width=\textwidth]{img/rt_\picvisi{_}.pdf}
        \end{subfigure}\vspace{.6ex}
        \begin{subfigure}[t]{\textwidth}
            \includegraphics[width=\textwidth]{img/rt_\picvisii{_}.pdf}
        \end{subfigure}\vspace{.6ex}
        \begin{subfigure}[t]{\textwidth}
            \includegraphics[width=\textwidth]{img/rt_\picvisiii{_}.pdf}
        \end{subfigure}\vspace{.6ex}
        \begin{subfigure}[t]{\textwidth}
            \includegraphics[width=\textwidth]{img/rt_\picvisiv{_}.pdf}
        \end{subfigure}\vspace{.6ex}
        \caption{\cite{bastani2018roadtracer}}
        \label{fig_qualitative_1st}
    \end{subfigure}
    \begin{subfigure}[t]{0.093\textwidth}
        \begin{subfigure}[t]{\textwidth}
            \includegraphics[width=\textwidth]{img/vc_\picvisi{_}.pdf}
        \end{subfigure}\vspace{.6ex}
        \begin{subfigure}[t]{\textwidth}
            \includegraphics[width=\textwidth]{img/vc_\picvisii{_}.pdf}
        \end{subfigure}\vspace{.6ex}
        \begin{subfigure}[t]{\textwidth}
            \includegraphics[width=\textwidth]{img/vc_\picvisiii{_}.pdf}
        \end{subfigure}\vspace{.6ex}
        \begin{subfigure}[t]{\textwidth}
            \includegraphics[width=\textwidth]{img/vc_\picvisiv{_}.pdf}
        \end{subfigure}\vspace{.6ex}
        \caption{\cite{tan2020vecroad}}
        \label{fig_qualitative_1st}
    \end{subfigure}
    \begin{subfigure}[t]{0.093\textwidth}
        \begin{subfigure}[t]{\textwidth}
            \includegraphics[width=\textwidth]{img/cb_\picvisi{_}.pdf}
        \end{subfigure}\vspace{.6ex}
        \begin{subfigure}[t]{\textwidth}
            \includegraphics[width=\textwidth]{img/cb_\picvisii{_}.pdf}
        \end{subfigure}\vspace{.6ex}
        \begin{subfigure}[t]{\textwidth}
            \includegraphics[width=\textwidth]{img/cb_\picvisiii{_}.pdf}
        \end{subfigure}\vspace{.6ex}
        \begin{subfigure}[t]{\textwidth}
            \includegraphics[width=\textwidth]{img/cb_\picvisiv{_}.pdf}
        \end{subfigure}\vspace{.6ex}
        \caption{\cite{liang2019convolutional}}
        \label{fig_qualitative_1st}
    \end{subfigure}
    \begin{subfigure}[t]{0.093\textwidth}
        \begin{subfigure}[t]{\textwidth}
            \includegraphics[width=\textwidth]{img/dm_\picvisi{_}.pdf}
        \end{subfigure}\vspace{.6ex}
        \begin{subfigure}[t]{\textwidth}
            \includegraphics[width=\textwidth]{img/dm_\picvisii{_}.pdf}
        \end{subfigure}\vspace{.6ex}
        \begin{subfigure}[t]{\textwidth}
            \includegraphics[width=\textwidth]{img/dm_\picvisiii{_}.pdf}
        \end{subfigure}\vspace{.6ex}
        \begin{subfigure}[t]{\textwidth}
            \includegraphics[width=\textwidth]{img/dm_\picvisiv{_}.pdf}
        \end{subfigure}\vspace{.6ex}
        \caption{\cite{homayounfar2019dagmapper}}
        \label{fig_qualitative_1st}
    \end{subfigure}
    \begin{subfigure}[t]{0.093\textwidth}
        \begin{subfigure}[t]{\textwidth}
            \includegraphics[width=\textwidth]{img/icurb_\picvisi{_}.pdf}
        \end{subfigure}\vspace{.6ex}
        \begin{subfigure}[t]{\textwidth}
            \includegraphics[width=\textwidth]{img/icurb_\picvisii{_}.pdf}
        \end{subfigure}\vspace{.6ex}
        \begin{subfigure}[t]{\textwidth}
            \includegraphics[width=\textwidth]{img/icurb_\picvisiii{_}.pdf}
        \end{subfigure}\vspace{.6ex}
        \begin{subfigure}[t]{\textwidth}
            \includegraphics[width=\textwidth]{img/icurb_\picvisiv{_}.pdf}
        \end{subfigure}\vspace{.6ex}
        \caption{\cite{zhxu2021icurb}}
        \label{fig_qualitative_1st}
    \end{subfigure}
    \begin{subfigure}[t]{0.093\textwidth}
        \begin{subfigure}[t]{\textwidth}
            \includegraphics[width=\textwidth]{img/ei_\picvisi{_}.pdf}
        \end{subfigure}\vspace{.6ex}
        \begin{subfigure}[t]{\textwidth}
            \includegraphics[width=\textwidth]{img/ei_\picvisii{_}.pdf}
        \end{subfigure}\vspace{.6ex}
        \begin{subfigure}[t]{\textwidth}
            \includegraphics[width=\textwidth]{img/ei_\picvisiii{_}.pdf}
        \end{subfigure}\vspace{.6ex}
        \begin{subfigure}[t]{\textwidth}
            \includegraphics[width=\textwidth]{img/ei_\picvisiv{_}.pdf}
        \end{subfigure}\vspace{.6ex}
        \caption{Ours}
        \label{fig_qualitative_1st}
    \end{subfigure}
    \caption{Qualitative demonstrations. Each row shows an example. The first column is the ground truth (cyan lines); column (b) to (d) are segmentation-based baselines (green lines); column (e) to (i) are graph-based baselines and the last column shows the results of the new proposed imitation-learning-based baseline (orange lines for edges and yellow nodes for vertices). For better visualization, the lines are drawn in a thicker width while they are actually one-pixel width. The figure is best viewed in color. Please zoom in for details.}
    \label{fig_qualitative}
\end{figure*}
\begin{table*}[th] 
\setlength{\abovecaptionskip}{0pt} 
\setlength{\belowcaptionskip}{0pt} 
\renewcommand\arraystretch{1.0} 
\renewcommand\tabcolsep{8.5pt} 
\centering 
\begin{threeparttable}
\caption{The quantitative comparative results. The best results are highlighted in bold font. For all the metrics, larger values indicate better performance.} 
\begin{tabular}{@{}c c c c c c c c c c c c c c c c@{}}
\toprule
\multirow{3}{*}{Methods}& \multicolumn{3}{c}{Precision $\uparrow$} & \multicolumn{3}{c}{Recall $\uparrow$} & \multicolumn{3}{c}{F1-score $\uparrow$}& \multirow{3}{*}{Naive $\uparrow$}& \multirow{3}{*}{APLS $\uparrow$} & \multirow{3}{*}{ECM $\uparrow$} \\ 
\cmidrule(l){2-4} \cmidrule(l){5-7} \cmidrule(l){8-10} 
&   2.0 &  5.0 &  10.0 &   2.0 &  5.0 &  10.0
&    2.0 &  5.0 &  10.0 & \\
\midrule
Naive baseline & 0.607&\textbf{0.890}&0.928&0.505&0.736&0.768&0.533&0.778&0.811& 0.698& 0.577& 0.550 \\
Deeproadmapper \cite{mattyus2017deeproadmapper} & 0.578&0.854&0.898&0.475&0.694&0.725&0.505&0.740&0.775 & 0.719 & 0.615 & 0.595 \\ 
OrientationRefine \cite{batra2019improved} &\textbf{0.620}&0.878&0.913&\textbf{0.602}&\textbf{0.850}&\textbf{0.884}&\textbf{0.605}&\textbf{0.855}&\textbf{0.888}& 0.797& 0.750& 0.756\\ 
\midrule
RoadTracer \cite{bastani2018roadtracer} & 0.391&0.707&0.791&0.416&0.743&0.821&0.399&0.718&0.798& 0.869& 0.739& 0.824
\\ 
VecRoad \cite{tan2020vecroad}&0.461&0.769&0.854&0.459&0.752&0.830&0.458&0.756&0.837& 0.883& 0.756& 0.846
\\ 
ConvBoundary \cite{liang2019convolutional} &  0.510 &0.845&\textbf{0.934}& 0.455&0.692&0.752& 0.465&0.737&0.805&\textbf{0.958}& 0.671& 0.786
\\ 
DAGMapper \cite{homayounfar2019dagmapper} & 0.407 & 0.751 & 0.868 & 0.353 & 0.649 & 0.747 & 0.371 & 0.684 & 0.787 & 0.896 & 0.679 & 0.758 
\\ 
iCurb \cite{zhxu2021icurb} & 0.550&0.833&0.890&0.538&0.815&0.873&0.542&0.820&0.877&0.910 & \textbf{0.826} & 0.889 \\
\midrule 
Enhanced-iCurb & 0.560&0.839&0.894&0.542&0.811&0.864&0.549&0.821&0.874&0.925& 0.822& \textbf{0.893}
 \\

\bottomrule 
\label{tab_comparative}
\end{tabular} 
\end{threeparttable}
\end{table*}

\subsection{The Proposed imitation-learning-based baseline}
We propose a new imitation-learning-based solution based on our previous work \textit{iCurb} \cite{zhxu2021icurb}, and the new solution is named as \textit{enhanced-iCurb}. To the best of our knowledge, \textit{iCurb} is the first work to solve the line-shaped object detection from the perspective of imitation learning. In \cite{zhxu2021icurb}, a DAgger-based solution is proposed. For each patch, the solution runs a round of restricted exploration as well as $N$ rounds of free exploration, and aggregates the training dataset using the generated samples. Let $\pi^*$ denote the expert policy and $\hat{\pi}$ denote the learner policy. The task is to learn the $\hat{\pi}$ to mimic $\pi^*$. Let $\hat{v}_t$ and $v_t^*$ respectively denote the actions produced by $\hat{\pi}$ and $\pi^*$ at time $t$. During the sampling period, $v_t$ is used to denote the vertex to update the graph. 

With the obtained prediction $\hat{v}_t$, we set the closest pixel of the ground-truth road boundary to $\hat{v}_t$ as the label $v_t^*$ to train \textit{iCurb}. This algorithm generates the label on-the-fly and can ensure \textit{iCurb} not be affected by the \textit{teacher-forcing} problem. However, $v_t^*$ heavily relies on $\hat{v}_t$ and does not have a unique value, thus $\hat{\pi}$ may make unpredictable actions and may converge to a sub-optimal policy. In our experiments, for example, we find that the edge length of the road boundary predicted by \textit{iCurb} is almost random sometimes. It requires careful parameter tuning to prevent this. To make the training process more stable and predictable, the orientation map $M_O$ is leveraged to calculate $v_t^*$ and $v_t^*$ has a unique value. We first obtain the radian $r_{t-1}$ of the previous vertex $v_{t-1}$, then we find the first vertex of the ground-truth road boundary in $M_O$ whose radian value has large enough difference with $r_{t-1}$, and this vertex is used as $v_t^*$ to train \textit{iCurb}. The new algorithm does not rely on $\hat{v}_t$ so that it guarantees unique $v_t^*$, which improves the quality of the final graph. So the label $v_t^*$ generated by \textit{enhanced-iCurb} is more reasonable and effective.

\section{Experimental Results and Discussions}
\subsection{Experimental setup}
We conduct experiments on a PC with an i7-8700K CPU, an NVIDIA GTX1080Ti GPU, and an RTX3090 GPU. 
The checkpoint with the best performance on the validation set is selected for inference. 
To measure the efficiency, we record the training time cost as well as the inference time cost of each baseline, and report their average time cost on each patch. For graph-based baselines, the ground-truth initial vertices added with Gaussian noise are used to start the iterative graph generation process. Considering the trade-off between efficiency and effectiveness, the number of rounds of the free exploration for both \textit{iCurb} and \textit{enhanced-iCurb} is set to 1.

\subsection{Evaluation results}
The comparative results are shown in Tab. \ref{tab_comparative}. Some qualitative demonstrations are shown in Fig. \ref{fig_qualitative}. The average processing time for efficiency evaluation is reported in Tab. \ref{tab_time}.  
As segmentation-based baselines directly optimize on pixel values, they tend to present good F1-scores, even the naive baseline. Segmentation-based baselines take relatively less time than graph-based baselines for training since they do not require an iterative process. However, these baselines do not have satisfactory performance on connectivity, because the spatial information of the patch image cannot be fully leveraged. Compared with the \textit{Naive Baseline}, \textit{DeepRoadMapper} can propose candidate connections to correct some disconnection and enhance the connectivity to some extent. But it is still restricted by the accuracy of the semantic segmentation. Once the segmentation network gives poor results, \textit{DeepRoadMapper} cannot effectively improve the final performance. Moreover, generating correction candidates increases the time cost. \textit{OrientationRefine} iteratively refines the obtained segmentation map in pixel-level with an extra network, which can obtain more concise correction results. Thus \textit{OrientationRefine} achieves good connectivity and outperforms all the other methods on F1-Score. 
\begin{table}[t] 
\setlength{\abovecaptionskip}{0pt} 
\setlength{\belowcaptionskip}{0pt} 
\renewcommand\arraystretch{1.0} 
\renewcommand\tabcolsep{16pt} 
\centering 
\begin{threeparttable}
\caption{The time consumption of the baselines. We report the average time taken to process a single image.} 
\begin{tabular}{c c c}
\toprule
Methods & Training & Inference \\
\midrule
Naive baseline & 0.45 s/image&0.33 s/image \\ 
Deeproadmapper\cite{mattyus2017deeproadmapper}&1.32 s/image&1.57 s/image\\ 
OrientationRefine\cite{batra2019improved}&1.14 s/image&0.96 s/image \\ 
\midrule
RoadTracer\cite{bastani2018roadtracer}&6.50 s/image &4.92 s/image \\ 
VecRoad\cite{tan2020vecroad}&19.89 s/image&14.17 s/image\\ 
ConvBoundary\cite{liang2019convolutional} & 17.47 s/image &  25.17 s/image\\ 
DAGMapper\cite{homayounfar2019dagmapper} & 8.19 s/image &  4.92 s/image\\ 
iCurb\cite{zhxu2021icurb} & 6.74 s/image &  3.11 s/image\\
\midrule 
Enhanced-iCurb & 7.25 s/image & 2.38 s/image\\
\bottomrule 
\label{tab_time}
\end{tabular} 
\end{threeparttable}
\end{table}

Instead of working on pixels, graph-based baselines generate the graph of road boundaries directly.
Thus, they perform well on connectivity. However, due to the sequential process for graph generation, this kind of methods exhibit relatively lower pixel-level accuracy when the patch image has complicated scenarios.
The training process of \textit{RoadTracer} is fast since it is a small network. But the performance is inferior to others. \textit{VecRoad} follows the similar idea of \textit{RoadTracer} but the network backbone is replaced with a more powerful one, so its performance is improved but the training efficiency is degraded. Even though \textit{ConvBoundary} and \textit{DagMapper} present good results with point-cloud maps on their original tasks, they do not present satisfactory performance on this task (i.e., road-boundary detection using aerial images). 
\textit{iCurb} provides a solution to this task using imitation learning. It gives better F1-Score and connectivity thanks to the DAgger-based training strategy. \textit{Enhanced-iCurb} utilizes different algorithms to generate the training label and update the graph, which is more stable and predictable. It has good performance on all metrics and qualitative visualizations, thus the superiority of our new proposed method is demonstrated.

\section{Conclusions and Future Work}
In this paper, we proposed a publicly available benchmark dataset named \textit{Topo-boundary} for topological road-boundary detection using BEV aerial images. \textit{Topo-boundary} has 25,295 patches. Each patch consists of a 4-channel aerial image and 8 labels for different deep-learning tasks. We also designed a new metric for better connectivity evaluation. It was employed to compare 9 baselines 
together with some metrics used in the past works. 
The dataset and our implemented code for the baselines were publicly available on our project page. 
In the future, we plan to assign the prediction difficulty scores (i.e., smaller values for easy cases, and larger values for hard cases) to further label the dataset, so that the networks can be trained to be well adapted to various scenarios. Moreover, the patches removed in this paper will be considered for real-world applications.

\bibliographystyle{IEEEtran}
\bibliography{mybib}

\begin{thebibliography}{10}
\providecommand{\url}[1]{#1}
\csname url@samestyle\endcsname
\providecommand{\newblock}{\relax}
\providecommand{\bibinfo}[2]{#2}
\providecommand{\BIBentrySTDinterwordspacing}{\spaceskip=0pt\relax}
\providecommand{\BIBentryALTinterwordstretchfactor}{4}
\providecommand{\BIBentryALTinterwordspacing}{\spaceskip=\fontdimen2\font plus
\BIBentryALTinterwordstretchfactor\fontdimen3\font minus
  \fontdimen4\font\relax}
\providecommand{\BIBforeignlanguage}[2]{{%
\expandafter\ifx\csname l@#1\endcsname\relax
\typeout{** WARNING: IEEEtran.bst: No hyphenation pattern has been}%
\typeout{** loaded for the language `#1'. Using the pattern for}%
\typeout{** the default language instead.}%
\else
\language=\csname l@#1\endcsname
\fi
#2}}
\providecommand{\BIBdecl}{\relax}
\BIBdecl

\bibitem{lu2020real}
X.~Lu, Y.~Ai, and B.~Tian, ``Real-time mine road boundary detection and
  tracking for autonomous truck,'' \emph{Sensors}, vol.~20, no.~4, p. 1121,
  2020.

\bibitem{zhu2015real}
X.~Zhu, M.~Gao, and S.~Li, ``A real-time road boundary detection algorithm
  based on driverless cars,'' in \emph{2015 4th National Conference on
  Electrical, Electronics and Computer Engineering}.\hskip 1em plus 0.5em minus
  0.4em\relax Atlantis Press, 2015, pp. 843--848.

\bibitem{sun20193d}
P.~Sun, X.~Zhao, Z.~Xu, R.~Wang, and H.~Min, ``A 3d lidar data-based dedicated
  road boundary detection algorithm for autonomous vehicles,'' \emph{IEEE
  Access}, vol.~7, pp. 29\,623--29\,638, 2019.

\bibitem{wang2019self}
H.~Wang, Y.~Sun, and M.~Liu, ``Self-supervised drivable area and road anomaly
  segmentation using rgb-d data for robotic wheelchairs,'' \emph{IEEE Robotics
  and Automation Letters}, vol.~4, no.~4, pp. 4386--4393, 2019.

\bibitem{mattyus2017deeproadmapper}
G.~M{\'a}ttyus, W.~Luo, and R.~Urtasun, ``Deeproadmapper: Extracting road
  topology from aerial images,'' in \emph{Proceedings of the IEEE International
  Conference on Computer Vision}, 2017, pp. 3438--3446.

\bibitem{batra2019improved}
A.~Batra, S.~Singh, G.~Pang, S.~Basu, C.~Jawahar, and M.~Paluri, ``Improved
  road connectivity by joint learning of orientation and segmentation,'' in
  \emph{Proceedings of the IEEE Conference on Computer Vision and Pattern
  Recognition}, 2019, pp. 10\,385--10\,393.

\bibitem{bastani2018roadtracer}
F.~Bastani, S.~He, S.~Abbar, M.~Alizadeh, H.~Balakrishnan, S.~Chawla,
  S.~Madden, and D.~DeWitt, ``Roadtracer: Automatic extraction of road networks
  from aerial images,'' in \emph{Proceedings of the IEEE Conference on Computer
  Vision and Pattern Recognition}, 2018, pp. 4720--4728.

\bibitem{tan2020vecroad}
Y.-Q. Tan, S.-H. Gao, X.-Y. Li, M.-M. Cheng, and B.~Ren, ``Vecroad: Point-based
  iterative graph exploration for road graphs extraction,'' in
  \emph{Proceedings of the IEEE/CVF Conference on Computer Vision and Pattern
  Recognition}, 2020, pp. 8910--8918.

\bibitem{homayounfar2019dagmapper}
N.~Homayounfar, W.-C. Ma, J.~Liang, X.~Wu, J.~Fan, and R.~Urtasun, ``Dagmapper:
  Learning to map by discovering lane topology,'' in \emph{Proceedings of the
  IEEE International Conference on Computer Vision}, 2019, pp. 2911--2920.

\bibitem{belli2019image}
D.~Belli and T.~Kipf, ``Image-conditioned graph generation for road network
  extraction,'' \emph{NeurIPS 2019 workshop on Graph Representation Learning},
  2019.

\bibitem{zhxu2021icurb}
Z.~Xu, Y.~Sun, and M.~Liu, ``icurb: Imitation learning-based detection of road
  curbs using aerial images for autonomous driving,'' \emph{IEEE Robotics and
  Automation Letters}, vol.~6, no.~2, pp. 1097--1104, 2021.

\bibitem{nyc_dataset}
N.~O. Department~of Information Technology \& Telecommunications~(DoITT),
  ``{NYC-Planimetrics Database},''
  \url{https://github.com/CityOfNewYork/nyc-planimetrics}, 2019.

\bibitem{van2018spacenet}
A.~Van~Etten, D.~Lindenbaum, and T.~M. Bacastow, ``Spacenet: A remote sensing
  dataset and challenge series,'' \emph{arXiv preprint arXiv:1807.01232}, 2018.

\bibitem{liang2019convolutional}
J.~Liang, N.~Homayounfar, W.-C. Ma, S.~Wang, and R.~Urtasun, ``Convolutional
  recurrent network for road boundary extraction,'' in \emph{Proceedings of the
  IEEE Conference on Computer Vision and Pattern Recognition}, 2019, pp.
  9512--9521.

\bibitem{mnih2010learning}
V.~Mnih and G.~E. Hinton, ``Learning to detect roads in high-resolution aerial
  images,'' in \emph{European Conference on Computer Vision}.\hskip 1em plus
  0.5em minus 0.4em\relax Springer, 2010, pp. 210--223.

\bibitem{homayounfar2018hierarchical}
N.~Homayounfar, W.-C. Ma, S.~Kowshika~Lakshmikanth, and R.~Urtasun,
  ``Hierarchical recurrent attention networks for structured online maps,'' in
  \emph{Proceedings of the IEEE Conference on Computer Vision and Pattern
  Recognition}, 2018, pp. 3417--3426.

\bibitem{li2018polymapper}
Z.~Li, J.~D. Wegner, and A.~Lucchi, ``Topological map extraction from overhead
  images,'' in \emph{Proceedings of the IEEE International Conference on
  Computer Vision}, 2019, pp. 1715--1724.

\bibitem{wegner2013higher}
J.~D. Wegner, J.~A. Montoya-Zegarra, and K.~Schindler, ``A higher-order crf
  model for road network extraction,'' in \emph{Proceedings of the IEEE
  Conference on Computer Vision and Pattern Recognition}, 2013, pp. 1698--1705.

\bibitem{haklay2008openstreetmap}
M.~Haklay and P.~Weber, ``Openstreetmap: User-generated street maps,''
  \emph{IEEE Pervasive computing}, vol.~7, no.~4, pp. 12--18, 2008.

\bibitem{ranzato2015sequence}
M.~Ranzato, S.~Chopra, M.~Auli, and W.~Zaremba, ``Sequence level training with
  recurrent neural networks,'' \emph{arXiv preprint arXiv:1511.06732}, 2015.

\bibitem{liang2018end}
J.~Liang and R.~Urtasun, ``End-to-end deep structured models for drawing
  crosswalks,'' in \emph{Proceedings of the European Conference on Computer
  Vision (ECCV)}, 2018, pp. 396--412.

\bibitem{huttenlocher1993comparing}
D.~P. Huttenlocher, G.~A. Klanderman, and W.~J. Rucklidge, ``Comparing images
  using the hausdorff distance,'' \emph{IEEE Transactions on pattern analysis
  and machine intelligence}, vol.~15, no.~9, pp. 850--863, 1993.

\bibitem{ronneberger2015u}
O.~Ronneberger, P.~Fischer, and T.~Brox, ``U-net: Convolutional networks for
  biomedical image segmentation,'' in \emph{International Conference on Medical
  image computing and computer-assisted intervention}.\hskip 1em plus 0.5em
  minus 0.4em\relax Springer, 2015, pp. 234--241.

\bibitem{gao2019res2net}
S.~Gao, M.-M. Cheng, K.~Zhao, X.-Y. Zhang, M.-H. Yang, and P.~H. Torr,
  ``Res2net: A new multi-scale backbone architecture,'' \emph{IEEE transactions
  on pattern analysis and machine intelligence}, 2019.

\bibitem{ross2011reduction}
S.~Ross, G.~Gordon, and D.~Bagnell, ``A reduction of imitation learning and
  structured prediction to no-regret online learning,'' in \emph{Proceedings of
  the fourteenth international conference on artificial intelligence and
  statistics}.\hskip 1em plus 0.5em minus 0.4em\relax JMLR Workshop and
  Conference Proceedings, 2011, pp. 627--635.

\end{thebibliography}

\end{document}